\documentclass{article} 
\usepackage{conf2023_conference,times}

\usepackage{amsmath,amsfonts,bm}

\def\eqref#1{equation~\ref{#1}}

\def\1{\bm{1}}

\def\rv{{\textnormal{v}}}

\def\vv{{\bm{v}}}

\DeclareMathAlphabet{\mathsfit}{\encodingdefault}{\sfdefault}{m}{sl}
\SetMathAlphabet{\mathsfit}{bold}{\encodingdefault}{\sfdefault}{bx}{n}

\DeclareMathOperator*{\argmax}{arg\,max}

\usepackage{hyperref}
\usepackage{url}

\usepackage{tikz}  
\usetikzlibrary{spy}
\usepackage{color, colortbl}
\usepackage{pgfplots}
\pgfplotsset{compat=newest}
\usepackage{multicol}
\usepackage{circuitikz}
\usepackage[nolist]{acronym}
\usepgfplotslibrary{groupplots}
\usetikzlibrary{shapes}
\usepackage{siunitx}  
\usepackage{graphicx}
\usepgfplotslibrary{patchplots}
\usepackage{amsmath,amssymb}
\usepackage{hhline}
\usepackage{subfig}
 \usepackage{float}
\usepackage{cite}
\usepackage{amsmath}
\usetikzlibrary{angles, quotes,calc,patterns}
\usepgfplotslibrary{fillbetween}
\definecolor{mittelblau}{RGB}{0, 126, 198}
\definecolor{violettblau}{cmyk}{0.9, 0.6, 0, 0}
\definecolor{rot}{RGB}{238, 28 35}
\definecolor{apfelgruen}{RGB}{140, 198, 62}
\definecolor{gelb}{RGB}{1, 221, 0}
\definecolor{orange}{RGB}{244, 111, 33}
\definecolor{pink}{RGB}{237, 0, 140}
\definecolor{lila}{RGB}{128, 10, 145}
\definecolor{hellgrau}{RGB}{224, 224, 224}
\definecolor{mittelgrau}{RGB}{128, 128, 128}
\definecolor{dunkelgrau}{RGB}{80,80,80}
\definecolor{anthrazit}{RGB}{19, 31, 31}
\definecolor{darkgreen}{RGB}{0.125,0.5,0.169}
\usepackage{booktabs}

\usepackage{ifthen}
\usepackage{algorithm}
\usepackage[noend]{algpseudocode}
\usetikzlibrary{matrix,calc}

\usepackage{wrapfig}

\usepackage[colorinlistoftodos]{todonotes}
\newcommand{\sysname}{RadioBench}

\title{Benchmarking Learnt Radio Localisation\\ under Distribution Shift}

\author{Max Arnold\thanks{Correspondence to \texttt{maximilian.wolfgang.arnold@gmail.com}} \\
Bell Labs \\
\And
Mo Alloulah \\
Bell Labs \\
}

\begin{document}
\renewcommand{\vec}[1]{\mathbf{#1}}
\newcommand{\vecs}[1]{\boldsymbol{#1}}

\newcommand{\av}{\vec{a}}
\newcommand{\bv}{\vec{b}}
\newcommand{\cv}{\vec{c}}
\newcommand{\dv}{\vec{d}}
\newcommand{\ev}{\vec{e}}
\newcommand{\fv}{\vec{f}}
\newcommand{\gv}{\vec{g}}
\newcommand{\hv}{\vec{h}}
\newcommand{\iv}{\vec{i}}
\newcommand{\jv}{\vec{j}}
\newcommand{\kv}{\vec{k}}
\newcommand{\lv}{\vec{l}}
\newcommand{\mv}{\vec{m}}
\newcommand{\nv}{\vec{n}}
\newcommand{\ov}{\vec{o}}
\newcommand{\pv}{\vec{p}}
\newcommand{\qv}{\vec{q}}
\renewcommand{\rv}{\vec{r}}
\newcommand{\sv}{\vec{s}}
\newcommand{\tv}{\vec{t}}
\newcommand{\uv}{\vec{u}}
\renewcommand{\vv}{\vec{v}}
\newcommand{\wv}{\vec{w}}
\newcommand{\xv}{\vec{x}}
\newcommand{\yv}{\vec{y}}
\newcommand{\zv}{\vec{z}}
\newcommand{\zerov}{\vec{0}}
\newcommand{\onev}{\vec{1}}
\newcommand{\alphav}{\vecs{\alpha}}
\newcommand{\betav}{\vecs{\beta}}
\newcommand{\gammav}{\vecs{\gamma}}
\newcommand{\lambdav}{\vecs{\lambda}}
\newcommand{\omegav}{\vecs{\omega}}
\newcommand{\sigmav}{\vecs{\sigma}}
\newcommand{\tauv}{\vecs{\tau}}

\newcommand{\Am}{\vec{A}}
\newcommand{\Bm}{\vec{B}}
\newcommand{\Cm}{\vec{C}}
\newcommand{\Dm}{\vec{D}}
\newcommand{\Em}{\vec{E}}
\newcommand{\Fm}{\vec{F}}
\newcommand{\Gm}{\vec{G}}
\newcommand{\Hm}{\vec{H}}
\newcommand{\Id}{\vec{I}}
\newcommand{\Jm}{\vec{J}}
\newcommand{\Km}{\vec{K}}
\newcommand{\Lm}{\vec{L}}
\newcommand{\Mm}{\vec{M}}
\newcommand{\Nm}{\vec{N}}
\newcommand{\Om}{\vec{O}}
\newcommand{\Pm}{\vec{P}}
\newcommand{\Qm}{\vec{Q}}
\newcommand{\Rm}{\vec{R}}
\newcommand{\Sm}{\vec{S}}
\newcommand{\Tm}{\vec{T}}
\newcommand{\Um}{\vec{U}}
\newcommand{\Vm}{\vec{V}}
\newcommand{\Wm}{\vec{W}}
\newcommand{\Xm}{\vec{X}}
\newcommand{\Ym}{\vec{Y}}
\newcommand{\Zm}{\vec{Z}}
\newcommand{\Lambdam}{\vecs{\Lambda}}
\newcommand{\Pim}{\vecs{\Pi}}

\begin{multicols}{2}
\begin{acronym}[WSSUS]
    \acro{3GPP}{3rd Generation Partnership Project}
    \acro{5G}{fifth-generation}
    \acro{ADC}{analog to digital converter}
    \acro{AoA}{angle-of-arrival}
    \acro{AE}{auto-encoder}
    \acro{AutoML}{automated machine learning}
    \acro{AFE}{analog front end}
    \acro{AGC}{automatic gain control}
    \acro{AGV}{automated guided vehicle}
    \acro{AMP}{approximate message passing}
    \acro{API}{Application Programming Interface}
    \acro{AWGN}{additive white Gaussian noise}

    \acro{BER}{bit error rate}
    \acro{BB}{baseband}
    \acro{bpcu}{bits per channel use}
    \acro{BP}{belief propagation}
    \acro{BPSK}{binary phase shift keying}
    \acro{BS}{base station}

    \acro{CB}{codebook}
    \acro{CC}{channel-charting}
    \acro{CT}{continuity}    
    \acro{CDF}{cumulative distribution function}
    \acro{CFO}{carrier frequency offset}
    \acro{CoSaMP}{compressive sampling matching pursuit}
    \acro{CP}{cyclic prefix}
    \acro{CS}{compressive sensing} 
    \acro{CSI}{channel state information}
    \acro{CNN}{convolutional neural network}

    \acro{DA}{domain adaptation}
    \acro{DAC}{digital-analog-converter}
    \acro{DC}{direct current}
    \acro{DE}{distance error}
    \acro{DeepL}{deep-learning}
    \acro{DoF}{degree-of-freedom}
    \acro{DFT}{discrete Fourier transformation}
    \acro{DL}{deep learning}
    \acro{DS}{delay spread}
    \acro{DSP}{digital signal processing}
	\acro{DNN}{deep neural network}
    \acro{ECC}{error-correcting code}
    \acro{ENoB}{effective number of bits}
    \acro{ERP}{effective radiated power}
    \acro{EVM}{error vector magnitude}
    \acro{EVD}{eigenvector decomposition}
    \acro{FB}{feedback}
    \acro{FC}{fully connected}
    \acro{FDD}{frequency division duplexing}
    \acro{FDM}{frequency division multiplexing}
    \acro{FIR}{finite impulse response}
    \acro{FFT}{fast fourier transform}
    \acro{FT}{fine tuning}
    \acro{FPGA}{field programmable gate array}
    \acro{GAN}{generative adversarial network}
    \acro{GPIO}{general-purpose input/output}
    \acro{GPS}{global positioning system}
    \acro{GPSDO}{GPS disciplined oscillator}
    \acro{GPU}{graphical processing unit}
    \acro{HDF}{Hierarchical Data Format}
    \acro{HDD}{hard decision decoding}
    \acro{IC}{integrated circuit}
    \acro{ICI}{inter-carrier-interference}
    \acro{ISAC}{Integrated Sensing And Communication}
    \acro{I2C}{Inter-Integrated Circuit}
    \acro{ICSP}{in-circuit serial programming}
    \acro{IF}{intermediate frequency}
    \acro{i.i.d.}{independent and identically distributed}
    \acro{IIR}{infinite impulse response}
    \acro{IMU}{inertial measurement unit}
    \acro{IoT}{Internet of Things}
    \acro{IPS}{indoor positioning system}
    \acro{IR}{infrared}
    \acro{JSDM}{Joint Spatial Division and Multiplexing}
    \acro{LIDAR}{Light Detection And Ranging}
    \acro{LLR}{log-likelihood ratio}
    \acro{LP}{leakage precoder}
    \acro{LMMSE}{Linear Minimum Mean Square Error}
    \acro{LO}{local oscillator}
    \acro{LoS}{line of sight}
    \acro{LiDaR}{Light Detection and Ranging}
    \acro{LS}{least squares}
    \acro{LSTM}{long-term short-term memory}
    \acro{LTE}{Long Term Evolution}
    \acro{LTI}{linear time invariant}
    \acro{LTV}{linear time variant}
    \acro{KS}{kruskal-stress}
  
    \acro{MAP}{maximum a posteriori}
    \acro{MDE}{mean distance error}
    \acro{MDA}{mean distance accuracy}
    \acro{MEMS}{Micro-Electro-Mechanical Systems}
    \acro{MIMO}{multiple input multiple output}
    \acro{MISO}{multiple input single output}
    \acro{ML}{machine learning}
    \acro{MLE}{maximum likelihood estimator}
    \acro{mMIMO}{massive multiple input multiple output}
    \acro{MMSE}{minimum mean square error}
    \acro{M-MMSE}{multi-cell minimum mean square error}
    \acro{MR}{maximum ratio}
    \acro{MRC}{maximum ratio combining}
    \acro{MRP}{maximum ratio precoding}
    \acro{MRT}{maximum ratio transmission}
    \acro{MSE}{mean squared error}
    \acro{MQTT}{Message Queuing Telemetry Transport}
    \acro{MU}{multi-user}
    \acro{MUSIC}{Multiple Signal Classification}
    \acro{NF}{noise figure}
    \acro{NN}{Neural Network}
    \acro{NNI}{Neural Network Intelligence}
    \acro{NLoS}{non-line of sight}
    \acro{NND}{neural network decoding}
    \acro{NTP}{Network Time Protocol}
    \acro{NMSE}{normalized mean squared error}
    \acro{NU}{not-used}
    \acro{OFDM}{orthogonal frequency division multiplex}
    \acro{OMP}{orthogonal matching pursuit}
    \acro{OPS}{outdoor positioning system}
    \acro{OT}{optimal transport}
    \acro{PB}{passband}    
    \acro{PER}{periodogram}
    \acro{PCB}{printed circuit board}
    \acro{PDR}{pedestrian dead reckoning}
    \acro{PDF}{probability density function}
    \acro{PDP}{power-delay-profile}
    \acro{PLL}{phase-locked-loop}
    \acro{PO}{phase-only}
    \acro{PPS}{pulse per second}
    \acro{QPSK}{quadrature phase shift keying}
    \acro{QuaDRIGa}{Quasi Deterministic Radio Channel Generator}

    \acro{RADAR}{Radio Detection And Ranging}
    \acro{ReLU}{rectified linear unit}
    \acro{RF}{radio frequency}
    \acro{RMS-DS}{Root Mean Square - Delay Spread}
    \acro{RNN}{recurrent neuronal network}
    \acro{RSSI}{received signal strength indicator}
    \acro{R-ZF}{regularized zero-forcing}
    \acro{SDD}{soft decision decoding}
    \acro{SDR}{software defined radio}
    \acro{SE}{spectral efficiency}
    \acro{SFO}{sampling frequency offset}
    \acro{STO}{sampling time offset}
    \acro{SLAM}{Simultaneous Localization and Mapping}
    \acro{SGD}{stochastic gradient descent}
    \acro{SISO}{single input single output}
    \acro{SINR}{signal-to-interference-and-noise-ratio}
    \acro{SIR}{signal-to-interference-ratio}
    \acro{SLNR}{signal-to-leakage-and-noise ratio}
    \acro{SNR}{signal-to-noise-ratio}
    \acro{SP}{subspace}
    \acro{SQR}{signal-to-quantization-noise-ratio}
    \acro{SQNR}{signal-to-quantization-noise-ratio}
    \acro{SVD}{singular value decomposition}
    \acro{SU}{single-user}
    \acro{TDD}{time division duplexing}
    \acro{ToA}{time-of-arrival}
    \acro{TW}{trustworthiness}    
    \acro{TAoA}{time-and-angle-of-arrival}
    \acro{TDoA}{time-difference-of-arrival}
    \acro{TRIPS}{time-reversal IPS}
    \acro{UE}{user equipment}
    \acro{UL}{uplink}
    \acro{ULA}{uniform line array}
    \acro{URLLC}{ultra-reliable low-latency communication}
    \acro{US}{uncorrelated scattering}
    \acro{USRP}{universal software radio peripheral}
    \acro{UWB}{ultra-wideband}
    \acro{WiFi}{Wireless Fidelity}
    \acro{WSS}{wide sense stationary}
    \acro{WSSUS}{wide sense stationary uncorrelated scattering}

    \acro{ZF}{zero forcing}
\end{acronym}
\end{multicols}

\maketitle

\begin{abstract}
Deploying radio frequency (RF) localisation systems invariably entails non-trivial effort, particularly for the latest learning-based breeds. 
There has been little prior work on characterising and comparing how learnt localiser networks can be deployed in the field under real-world RF distribution shifts.
In this paper, we present \sysname: a suite of 8 learnt localiser nets from the state-of-the-art to study and benchmark their real-world deployability, utilising five novel industry-grade datasets. 
We train 10k models to analyse the inner workings of these learnt localiser nets and uncover their differing behaviours across three performance axes: (i) learning, (ii) proneness to distribution shift, and (iii) localisation. 
We use insights gained from this analysis to recommend best practices for the deployability of learning-based RF localisation under practical constraints.
\end{abstract}

\section{Introduction} \label{sec:intro}

Decades of of radio frequency (RF) localisation research have given us a variety of classic methods~\citep{patwari2005locating,gezici2005localization}.
Newer machine learning incarnations can enhance location estimation considerably~\citep{zanjani2022deep,karmanov2021wicluster}, albeit at the expense of proneness to distributional shift in wireless signals.
For example, models trained on signals from a warehouse environment may not work well in another different environment~\citep{arnold2018}.
If learnt localiser networks are to be productised and deployed, it is imperative that we robustify them.
To achieve real-world robustness, we need to understand (i) the performance nuances of learnt localisation models, (ii) when, how, and why do such models work, and (iii) when do they fail.

Robustness to distribution shift (i.e., out of distribution (OOD) generalisation) is an established line of enquiry in mainstream machine learning~\citep{gulrajani2020search,hendrycks2021many,koh2021wilds}.
However, there is little in the way of robustness investigations for learnt RF localisation.
Though lower dimensional than images, wireless signals are prone to acute variabilities stemming from environment- and/or system-dependent propagation conditions~\citep{tse2005fundamentals}, which are hard to control for.
Sidestepping this complexity, recent works have incorporated  environment-dependent priors (e.g., floorplans) in order to achieve robust learnt RF localisation in that environment ~\citep{karmanov2021wicluster,zanjani2022deep,ghazvinian2021modality}.

In this paper, we seek to understand the practical deployability of learnt RF localiser nets from first principles and without invoking extra robustifying priors.
To this end, we build~\sysname: a suite of RF localiser nets from the state-of-the-art.
We conduct a systematic comparative study on these localiser nets, utilising five novel industry-grade datasets. 
We analyse the inner workings of these localiser nets and uncover their differing behaviours across three performance axes: (i) learning, (ii) proneness to distribution shift, and (iii) localisation. 
Our contributions are:
\begin{itemize}
  \item We introduce \sysname: a benchmarking suite of RF localiser nets from the state-of-the-art, as well as a best-in-class classical probabilistic approach.
  \item We introduce 5 large-scale, industry-grade RF localisation datasets with differing characteristics that pertain to the study of wireless OOD robustness.
  \item We characterise and contrast the performance of 8 RF localiser methods, training in excess of 10k models in the process. These model configurations span: architecture, representation learning, and domain adaptation methods. 
  \item We find that representation learning and pretraining are most important for OOD robustness in a new RF environment, and that variants based on an autoencoder architecture are the best all-rounder models.
\end{itemize}

\section{Primer on RF Localisation} \label{sec:primer_rf_loc}

We consider a system of $M$ synchronised locators that listen for user devices, where each $m$th locator has known 3D position vector $\mathbf{u}_m = [x_m, y_m, z_m]$, and 3D 3$\times$3 orientation matrix $\mathbf{\Omega}_{m}$. 
Let $\mathbf{A}$ be the angle of arrival (AoA) matrix, $r$ the range calculated using time of arrival (ToA) and the speed of the light, then the position of a user device w.r.t. $m$th locator
\begin{equation}\label{eq:perfect}
\pv_m =  \mathbf{\Omega}_m \mathbf{A} \, r + \uv_m
\end{equation}
Because user devices and locators are not synchronised, range estimates are biased.
This can be compensated by using one locator as reference using time difference of arrival (TDoA). 
Typically, modern RF localisation relies on estimating the aforementioned two wireless propagation properties ToA and AoA, which together are abbreviated (TAoA).

\noindent \textbf{Challenge in rich scattering.} 
Considering a wireless channel between two radio transceivers, the baseband model of the channel impulse response is given by~\citep{tse2005fundamentals}
\begin{equation*}\label{eq:corr_fundamental}
h(k) = \sum_{p=1}^{P}  \sum_{\ell=0}^{L-1} a_{p,\ell} e^{j\left(2\pi f_c \tau_p +\phi_{p,\ell}\right)} \text{sinc}\left(k -\frac{\tau_{p,\ell}}{T_s}\right),\quad k=0,\dots,O-1
\end{equation*}
where $a_p\in \mathbb R_+$, $\phi_{p,\ell} \in \mathbb R$, $\tau_p\in \mathbb R_+$ are respectively the attenuation, phase, and propagation delay of the $p$th path and $\ell$th path cluster. 
Also $\text{sinc}(x)=\frac{\sin(\pi x)}{\pi x}$ is the normalised sinc function, $k$ is the discrete sampling time, and $O-1$ is the channel order.  

It is generally infeasibly to estimate the above parameters because they are underdetermined in practical implementations.
This is further compounded by environments with rich scattering (i.e., large $P$ and $L$).

\noindent \textbf{Upper bound.}
Eq.~\ref{eq:perfect} shows that the best performance can be theoretically achieved using perfect TAoA labels as input to a deep neural net. 
TAoA, however, are infeasible to measure as groundtruth per deployed environment because it would entail extensive and very expensive surveying campaigns.
Deployment surveys typically leverage laser measurements and tens of hours of calibration~\citep{scott2003user}.
Further, moving from a local coordinate system (i.e., per locator) to a global coordinate system for the environment requires models of that environment and the locator hardware. 
Therefore, we designate a TAoA-based localiser net as an upper bound on performance that is impractical to implement in the real-world under realistic deployment cost and overhead constraints. 

\section{Model Variants} \label{sec:model_varients}

\sysname~suite compiles all RF localiser net architectures reported in literature.
While all facilitate location estimation, these architectures operate on differing input formats, produce differing output formats, as well as deviate in their training details.
We believe~\sysname~to be the first effort to comprehensively catalogue and evaluate RF localiser nets in order to concretely establish and contrast their performances.
Appendix~\ref{sec:rf_loc_background} reviews RF localisation fundamentals and treats learnt localiser net variants in more detail.

\subsection{Architectures}
We evaluate four classes of RF localiser nets: supervised CNN~\citep{chen2017ConFI,arnold2019novel}, supervised residual net (ResNet) akin to vision ResNet~\citep{he2016deep}, unsupervised AutoEncoder (AE)~\citep{Jing2018}, and unsupervised channel charting (CC)~\citep{studer2018channel}.

\subsection{Input-Output (IO) formats}

The above architectures can ingest various representations of input wireless signals.
These are: (i) Channel state information (CSI) is the raw measurements obtained from transceiver chips, (ii) periodograms (PER) is CSI's 2D Fourier projection, (iii) a feature reduced version of (i) or (ii), and (iv) TAoA are the
physical propagation primitives that implicitly encode location (cf., Eq.~\ref{eq:perfect}), and are obtained via surveying the environment 
as discussed in Sec.~\ref{sec:primer_rf_loc}.

The above architectures can also produce multiple output representations.
These representations either encode location directly, or encapsulate it indirectly.
Specifically, output can be: (i) position estimate, (ii) TAoA primitives, or (iii) latent space that implicitly contains the location intrinsic space. 

Tab.~\ref{tab:model_variants} lists all valid architecture-IO configurations supported in~\sysname. 
Specifically for each method, Tab.~\ref{tab:model_variants} shows the effective mapping and its optimisation objective, which is minimisation for AE and maximisation for CC.
For further details around these methods, consult original literature.

\begin{table}[h]
\centering
\caption{RF localiser nets and their valid architecture, input-output configurations, and training objective implemented in~\sysname.}
\vspace{-0.2cm}
\scriptsize
\label{tab:model_variants}
    \begin{tabular}{rccc}
    \toprule
    \multicolumn{1}{c}{Detail}  & Supervised & Autoencoder (AE) & Channel chart (CC) \\
    \cmidrule(lr){1-4}
    Mapping &  \(\displaystyle  \mathbb {C}^{M} \to \mathbb {R} ^{3}\)            &    \(\displaystyle  \mathbb {C}^{M} \to \mathbb {R} ^{M'}\)                      &     \(\displaystyle  \mathbb {C}^{M} \to \mathbb {R} ^{2}\)                      \\
    Optimisation         &   \(\displaystyle \|\mathbf {p} _{n} - g (\mathbf {f} _{n})\|_{2}^{2}   \)      &  \(\displaystyle \|\mathbf {f} _{n} - g^{-1}(g (\mathbf {f} _{n}))\|_{2}^{2}   \)         &   \(\displaystyle \|{d\left(g(T_a),g(T_p)\right)-d\left(g(T_a),g(T_n)\right)}\| \)               \\
    Type         & ResNet     & CNN          & CNN              \\
    Input        & CSI/PER    & CSI/PER      & \(\displaystyle  f(\text{CSI}/\text{PER}) \)      \\
    Output       & Position/TAoA   & $M'$ Features     & Channel Chart   \\
    \cmidrule(lr){1-4}
    Configuration        & \parbox{4cm}{\centering CSI2Pos, PER2Pos\\ CSI2TAoA, PER2TAoA}    & CSI AE, PER AE     & CSI CC, PER CC   \\
    \bottomrule
    \end{tabular}
    \vspace{-0.5cm}
\end{table}

\subsection{Classical baseline}

A best-in-class probabilistic method is also used for benchmarking~\citep{henninger2022probabilistic}, which we designate as Classical.
Classical uses super-resolution techniques to estimate TAoA from CSI. 
These TAoAs are inputted to a maximum likelihood estimation (MLE) pipeline. 
Note that Classical results presented throughout paper are averaged per grid position for best-case analysis.

\section{Framework for Empirical OOD Robustness Analysis} \label{sec:analysis_framwork}

We introduce five large-scale, industry-grade datasets that enable us to empirically study the OOD robustness of wireless localiser nets.
We review the distribution shift mechanisms at play in these datasets.
We then discuss the distribution shift mitigation strategies we deem applicable to the model variants of Sec.~\ref{sec:model_varients}.

\begin{table}
  \centering
  \caption{Datasets utilised in evaluation and their configurations.}
  \label{Tab:datasets}
  \vspace{-0.4cm}
  \scriptsize
  \begin{tabular}{rrrrrrrrlr}
    \toprule
    \# & \multicolumn{1}{c}{Dataset}     & \multicolumn{1}{c}{points}   & \multicolumn{1}{c}{$f_c$ (GHz)}   & \multicolumn{1}{c}{Bandwidth (MHz)}   & \multicolumn{1}{c}{Subcarriers}   & \multicolumn{1}{c}{Antennae}  & \multicolumn{1}{c}{Locators}  & \multicolumn{1}{c}{Groundtruth$^{\ast}$$^{\dagger}$}   & \multicolumn{1}{c}{Area ($m^2$)}  \\[-0.0cm]
    \cmidrule(lr){2-10} 
    1 & Arena 1        & 52991         & 3.75          & 100               & 1630          &  8        & 6         & SLAM          & 134.532   \\
    2 & Arena 2        & 46266         & 3.75          & 100               & 1630          &  8        & 6         & SLAM          & 134.709   \\
    3 & Arena 3        &  2181         & 3.75          & 100               & 1630          &  8        & 6         & SLAM          &  57.188   \\
    4 & Industry 1     &  7990         & 3.75          & 100               & 1630          &  8        & 6         & Tachy         & 387.872   \\
    5 & Industry 2     &  5037         & 3.75          & 100               & 1630          &  8        & 6         & Tachy         & 103.469   \\
    \bottomrule
  \end{tabular}
  \raggedright
  \scriptsize{$^{\ast}$SLAM: obtained from simultaneous localization and mapping of mobile robot equipped with Lidar \\[-0.05cm] $^{\dagger}$Tachy: high-precision laser surveying device}
  \vspace{-0.4cm}
\end{table}
\subsection{Datasets}
\begin{figure}[htb]
    \centering
    \includegraphics{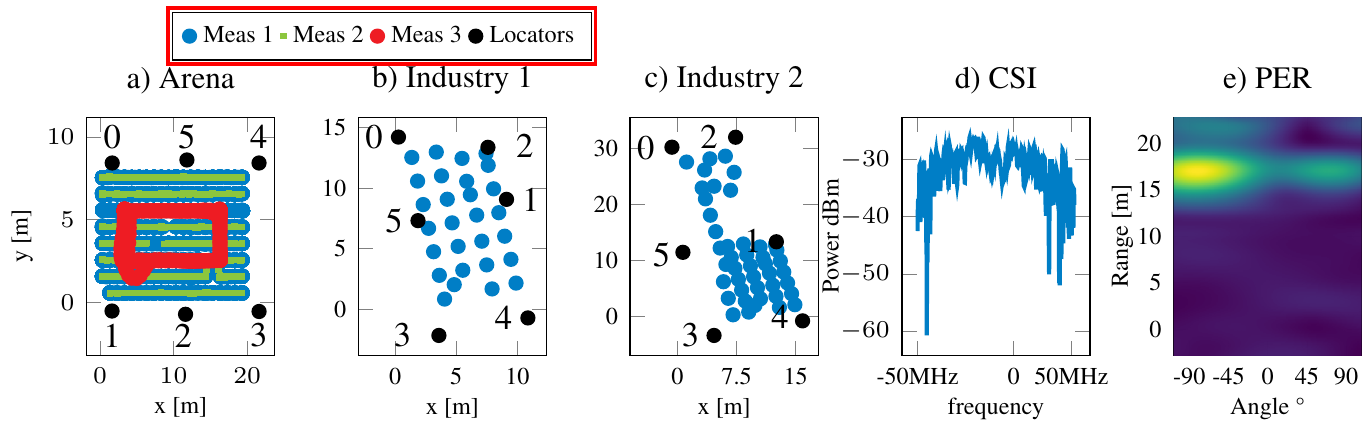}
    \vspace{-0.65cm}
    \caption{Measurement environments. a) Arena has three different measurement iterations, b) Typical Industrial environment, and c) Harsher industrial environment with rich scatterers. Data examples: d) CSI and e) PER.}
    \label{fig:lmu_experiments}
\end{figure}

We utilise a set of empirical industry-grade datasets to study the nuances of radio localiser net variants.
We summarise the setup and geometric configurations under which the radio measurement campaign was conducted.

Fig.~\ref{fig:lmu_experiments} depicts three physically distinct environments.
Within each, six locators listen to mobile users.
Each locater is equipped with a 3$\times$3 antenna array.
Locators are tightly synchronised using White Rabbit standard~\citep{eidson2002ieee}.
Mobile user devices regularly transmit pilot data known to the locators.
The locators receive user pilots and estimate their CSI. 
Fig.~\ref{fig:lmu_experiments}a shows 3 Arena 1 measurements in blue, green, and red.
These correspond to three data collection iterations.
Arena 1 and 2 (blue and green) cover the same area but are different due to hardware effects.
Arena 3 (red rectangle) corresponds to high-speed driving to simulate a dynamic environment.
Fig.~\ref{fig:lmu_experiments}b~\&~c correspond to two other industrial environments, with Industry 2 being particularly rich in scattering effects. 
Fig.~\ref{fig:lmu_experiments}d~\&~e depict two examples of the input formats discussed in Tab.~\ref{tab:model_variants}.

Tab.~\ref{Tab:datasets} summarises the configurations of our 5 novel datasets.

\subsection{Distribution shift mechanisms}

From the 5 datasets listed in Tab.~\ref{Tab:datasets}, we highlight the following mechanisms that result in distributional shift in RF signals.

\noindent \textbf{(1) Macro environment-induced.} 
Each of the three environments depicted in Fig.~\ref{fig:lmu_experiments} comes with its signature set of propagation conditions. 
These propagation conditions are largely a function of the geometry of the environment as well as the spatial configuration of reflective surfaces present within, e.g., metallic machinery, furniture, partition walls and their material composition, etc.
We designate environmental signatures as macro-level effects that shift the bulk of the distribution of RF signals.

\noindent \textbf{(2) Micro locator-induced.}
The 3D position and orientation of locators within the environment affect how they measure the statistics of user RF signals.
That is, a locator will also modulate the distribution of the RF signals it receives.
We designate locator signatures as micro-level effects that further shift the distribution of RF signals.

\noindent \textbf{(3) Micro scattering-induced.}
Dynamic activities within the environment induce scattering effects that modulate the distribution of the RF signals.
Example scatterers include moving robots and people.
We designate scattering as micro-level effects that further shift the distribution of RF signals.

\noindent \textbf{(4) Misc.}
For completeness, there are multiple other factors that affect the distribution of RF signals.
Examples include hardware- and frequency-dependent effects.
We, however, are mainly interested in shift mechanisms 1-3 in this work as captured by our 5 empirical datasets.

\subsection{Distribution shift mitigations} \label{sec:analysis_shift_mitigations}

There are a wide range of methods from the state-of-the-art that enhances robustness and generalisation on unseen distribution shifts.
However, the relative performance of these methods varies largely across modalities, datasets, and distribution shifts~\citep{hendrycks2021many,koh2021wilds,wiles2021}.
Further, there is little prior experience in adapting some of these concepts to the RF localiser net setting we study herein.

\noindent \textbf{Loss landscape.}
Not all models are created equal.
For all models described in Sec.~\ref{sec:model_varients}, we visualise a dataset's loss landscape using~\citep{li2018visualizing}.
This analysis is motivated by the observation that the landscape geometry affects generalisation dramatically~\citep{li2018visualizing}.
All things being equal, we would therefore favour model variants that exhibit flatter loss landscape geometry.
Our intuition is that a flatter loss landscape would readily support a weak form of generalisability.

\noindent \textbf{Fine-tuning.}
Fine-tuning is a consistent indicator of of the quality of zero-shot models~\citep{radford2021learning,wortsman2022robust}.
It is decoupled from some modality-specific recipes such as data augmentation.
Therefore, we employ simple universal fine-tuning for zero-shot model variants from Sec.~\ref{sec:model_varients} to gauge their relative robustness and generalisability to unseen distribution shifts.

\subsection{Learnability conditions}

We characterise various aspects around the learnability of the model variants described in Sec.~\ref{sec:model_varients}.

\noindent \textbf{Active label density.}
We investigate the required number of labels for validation loss convergence.
We employ active learning strategies to glean comparative insights on the learning behaviour of localiser model variants. 

\noindent \textbf{Latent space.}
Some model variants utilise a latent space that implicitly encodes location.
We investigate the resultant shift in the latent space as a function of macro and micro RF signal distribution shifts.

\noindent \textbf{Regression head protocol.}
For model variants with a latent space, we investigate the feasibility of a regressor head on top of a frozen backbone that is trained on a different dataset.
We intuit that if quality features have been learnt, their projection would still perform competitively w.r.t. regressing location estimates notwithstanding distribution shift.

\section{Experiments}\label{sec:experiments}

We evaluate 8 RF localiser nets.
We conduct a comprehensive analysis to quantify performance aspects around: (i) learnability, (ii) proneness to distribution shift, and (iii) localisation.
We use our 5 industry-grade datasets (cf., Tab.~\ref{Tab:datasets}) for all experiments.
We either use all or a subset of localiser model variants (cf., Tab.~\ref{tab:model_variants}) depending on suitability for a given experiment.
We begin by distilling our experimental findings into a concrete set of key takeaways.

As discussed in Sec.~\ref{sec:primer_rf_loc}, method TAoA2Pos in all analyses represents an upper bound on performance.
This is because in practical deployments, surveying groundtruth TAoAs is prohibitively expensive.

\subsection{Takeaways}

\begin{figure}[b]
   \centering
   \includegraphics{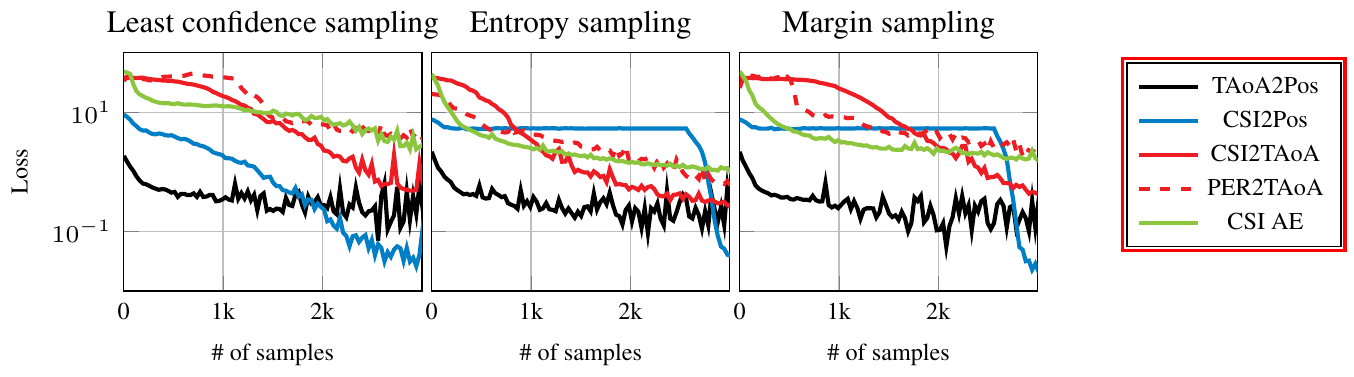}
   \vspace{-0.35cm}
   \caption{Active learning (AL) for model variants on Arena 1. AL criteria help models converge faster in required training samples. Required number of training samples is of the order of a dataset's spatial location sampling grid. Some model variants are qualitatively better than others from a learning standpoint.}
   \label{fig:active_learning}
\end{figure}

\begin{figure}[t] 
    \centering
    \includegraphics{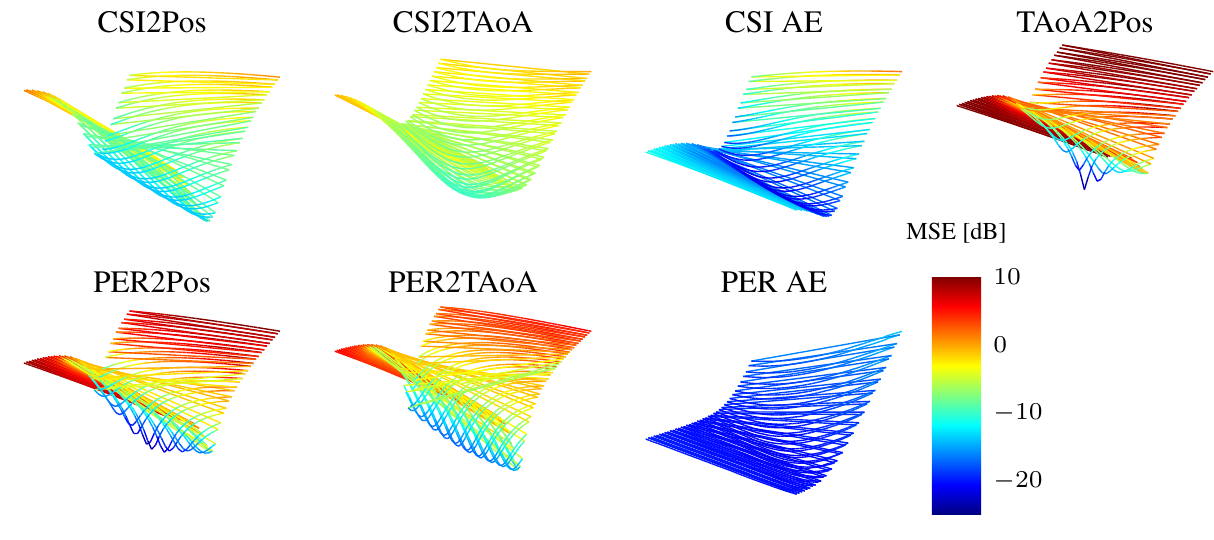}
    \vspace{-0.25cm}
    \caption{Loss landscapes of model variants on Arena 1. Loss landscapes are a proxy to inspecting the convex hull of model variants. The geometry of a model's loss landscape is indicative of its generalisability.}
    \label{fig:loss_landscapes}
\end{figure}

\noindent \textbf{1 -- Learnability: Under smart sample selection criterion, training samples of the order of the spatial grid suffice for convergence.}
We observe that models CSI AE, CSI2TAoA, and PER2TAoA converge after selecting a number samples (via active learning) comparable to the number of spatial grid locations.
Inspecting Fig.~\ref{fig:active_learning}, this happens after around 2.7k samples.
This is inline with Arena 1 dataset that has around 53k data points of which 2.7k are spatial grid locations (cf., Fig.~\ref{fig:lmu_experiments}a).

\noindent \textbf{2 -- Learnability: Model variants exhibit qualitative learning differences.} 
Methods that directly map to position (i.e., CSI2Pos and PER2Pos) are poor learners.
This phenomenon is most evident in Fig.~\ref{fig:active_learning}'s entropy and margin sampling losses where CSI2Pos shows a sudden drop around 2.7k samples, which hints at memorisation (i.e., fingerprinting of environment).
In contrast, methods that indirectly map to a local ambient space (i.e., CSI AE, CSI2TAoA, and PER2TAoA) do not exhibit such a waterfall effect in their learning loss.

\noindent \textbf{3 -- Shift: Model architecture and training details imbue a weak sense of generalisability by construction.} 
RF localiser nets based on an AE architecture (i.e., CSI \& PER AEs in Fig.~\ref{fig:loss_landscapes}) has the best zero-shot weak generalisability owing to flatter loss landscape~\citep{li2018visualizing} (cf., Sec.~\ref{sec:analysis_shift_mitigations}).
CSI \& PER AEs are agnostic to any position-dependent information (direct or indirect) and are trained only on a reconstruction loss. 
In contrast, methods that use information that encodes position during training (i.e., TAoA2Pos and Per2Pos) are at the opposite end of convexity steepness, necessitating more work for shift mitigation.

\begin{table}[h]
\centering
\caption{Zero-shot OOD performance. Performance reported in terms of median error for position and/or azimuth/elevation/range.}
\vspace{-0.25cm}
\label{tab:0shot_gen}
\begin{minipage}{.175\textwidth}
\centering
\scriptsize
    \begin{tabular}{c}
        \toprule
        Input        \\ 
        \midrule
        Output       \\ 
        \midrule
        Train $\rightarrow$ Test         \\ 
        \midrule
        \multicolumn{1}{l}{Arena 1 $\rightarrow$ Arena 1}      \\ 
        \multicolumn{1}{l}{Arena 1 $\rightarrow$ Arena 2}      \\ 
        \multicolumn{1}{l}{Arena 1 $\rightarrow$ Arena 3}      \\ 
        \multicolumn{1}{l}{Arena 1 $\rightarrow$ Industry 1}   \\ 
        \multicolumn{1}{l}{Arena 1 $\rightarrow$ Industry 2}   \\ 
        \bottomrule
    \end{tabular}
\end{minipage}%
\begin{minipage}{.31\textwidth}
\centering
\scriptsize
    \begin{tabular}{ccc}
        \toprule
        CSI                  & PER                  & TAoA               \\ 
        \cmidrule(lr){1-3}
        Position             & Position             & Position           \\ 
        \cmidrule(lr){1-3}
                             &                      &                    \\[+0.20cm] 
        \SI{0.07}{\metre}    &   \SI{0.03}{\metre}  &  \SI{0.03}{\metre} \\ 
        \SI{0.65}{\metre}    &   \SI{0.19}{\metre}  &  \SI{0.03}{\metre} \\ 
        \SI{1.09}{\metre}    &   \SI{0.62}{\metre}  &  \SI{0.11}{\metre} \\ 
        \SI{4.70}{\metre}    &   \SI{4.31}{\metre}  &  \SI{2.76}{\metre} \\ 
        \SI{8.44}{\metre}    &   \SI{7,83}{\metre}  &  \SI{6.01}{\metre} \\ 
        \bottomrule
    \end{tabular}
\end{minipage}%
\hspace{-0.35cm}
\begin{minipage}{.3\textwidth}
\centering
\scriptsize
    \begin{tabular}{cc}
        \toprule
        CSI                                                       & PER              \\ 
        \cmidrule(lr){1-2}
        TAoA                                                      & TAoA             \\ 
        \cmidrule(lr){1-2}
                                                                  &                  \\[+0.20cm] 
        \SI{0.03}{\degree}/\SI{0.02}{\degree}/\SI{0.01}{\metre}   &   \SI{0.05}{\degree}/\SI{0.02}{\degree}/\SI{0.007}{\metre}  \\ 
        \SI{0.31}{\degree}/\SI{0.12}{\degree}/\SI{0.04}{\metre}   &   \SI{0.05}{\degree}/\SI{0.02}{\degree}/\SI{0.007}{\metre}  \\ 
        \SI{0.75}{\degree}/\SI{0.37}{\degree}/\SI{0.09}{\metre}   &   \SI{0.17}{\degree}/\SI{0.07}{\degree}/\SI{0.03}{\metre}   \\ 
        \SI{5.73}{\degree}/\SI{1.46}{\degree}/\SI{0.35}{\metre}   &   \SI{0.49}{\degree}/\SI{0.18}{\degree}/\SI{0.07}{\metre}   \\ 
        \SI{3.49}{\degree}/\SI{1.89}{\degree}/\SI{0.97}{\metre}   &   \SI{5.88}{\degree}/\SI{1.54}{\degree}/\SI{0.50}{\metre}   \\ 
        \bottomrule
    \end{tabular}
\end{minipage}
\hspace{-0.065cm}
\end{table}

\noindent \textbf{4 -- Shift: Zero-shot OOD performance corroborates that not all models are created equal w.r.t. robustness.} 
The representation used for the mapping between the input and output within RF localiser nets has a large bearing on the network's zero-shot OOD robustness.
To examine this, we train all model variants on Arena 1 and test them on all datasets.
We conduct one-off calibration of the models' mean position and TAoA estimates to coarsely correct for distribution shifts.
Tab.~\ref{tab:0shot_gen} summarises the median performance obtained under this zero-shot domain adaptation setting.
We can see that models with direct position mapping significantly underperform their TAoA mapping counterparts.
This finding is another restatement of the observation in Takeaway 3 illustrated in Fig.~\ref{fig:loss_landscapes}.

\begin{figure}[hbt]
    \centering
    \vspace{-0.35cm}
    \includegraphics{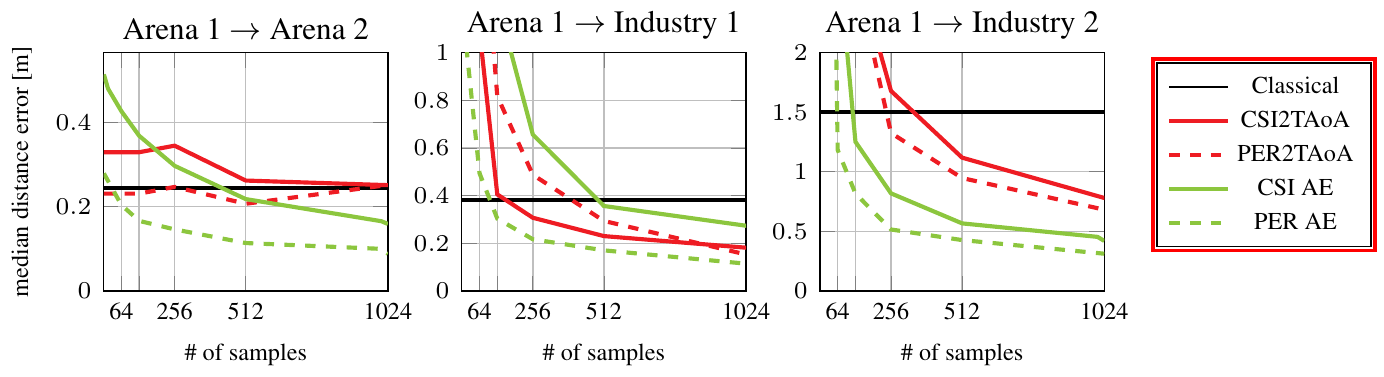}
    \vspace{-0.45cm}
    \caption{Performance of fine-tuned pretrained models.}
    \label{fig:perf_finetune}
     \vspace{-0.25cm}
\end{figure}

\noindent \textbf{5 -- Shift: Fine-tuning on few hundred labels achieves good domain adaptation performance across macro and micro  distribution shifts.}
Simple fine-tuning on a new domain is a robust means to significant performance boosting irrespective of the shift mechanism nature and magnitude.
Inspecting Fig.~\ref{fig:perf_finetune}, a fraction of the label density of the spatial grid is required, i.e., around 1/10th to 1/5th depending upon dataset and model variant.
AE variants (i.e., CSI \& PER AEs) are most effective.
When fine-tuned, pretrained localiser nets outperform the best classic baseline by at least 2$\times$ to 3$\times$ at the 50th $\%$ile error, again depending upon dataset and model variant.

\begin{figure}[htb]
  \centering
  \includegraphics{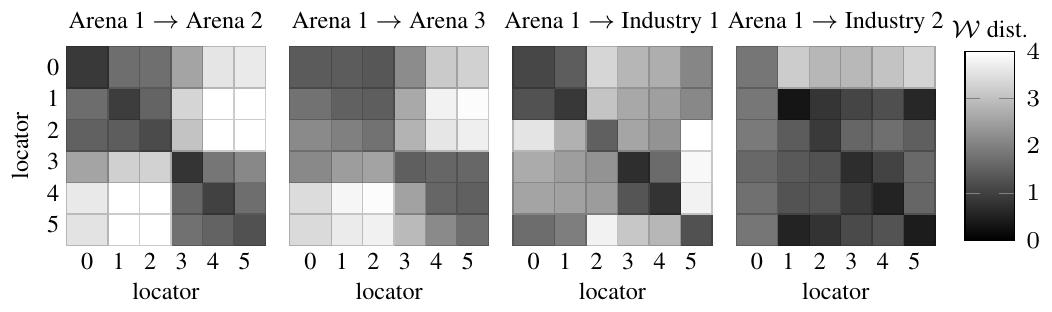}
  \vspace{-0.25cm}
  \caption{Wasserstein ($\mathcal{W}$) distances matrix that quantifies the amount of work needed to translate Arena 1 to other macro and micro domains for the CSI AE latent space. $\mathcal{W}$ is averaged across the spatial sampling grid.}
      \vspace{-0.25cm}
  \label{fig:wasserstein_matrix}
\end{figure}

\noindent \textbf{6 -- Shift: The latent space mirrors macro and micro distribution shifts.}
The high dimensionality of the AE latent space affords avenues for applying wider range of domain generalisation techniques.
To quantify the inter-dataset shifts seen in AE latent space, Fig.~\ref{fig:wasserstein_matrix} measures the Wasserstein distances~\citep{peyre2017computational} across our 5 datasets---relative to a model trained on Arena 1.
Locater-induced micro shifts cluster the latent spaces of Arena 1 $\rightarrow$ Arena 2-3 into two groups: 0-2 and 3-5.
This is in line with the physical nature of Arena configurations as groups 0-2 and 3-5 differ in their orientation for maximum spatial coverage (see Fig.~\ref{fig:lmu_experiments}a).
On Arena 1 $\rightarrow$ Industry 1-2, the picture is more nuanced.
For instance, Arena 1 $\rightarrow$ Industry 2 shows that locator 0, which is especially rich in scattering, exhibits larger distances to all other locators.

\begin{figure}[h]
   \centering
   \includegraphics{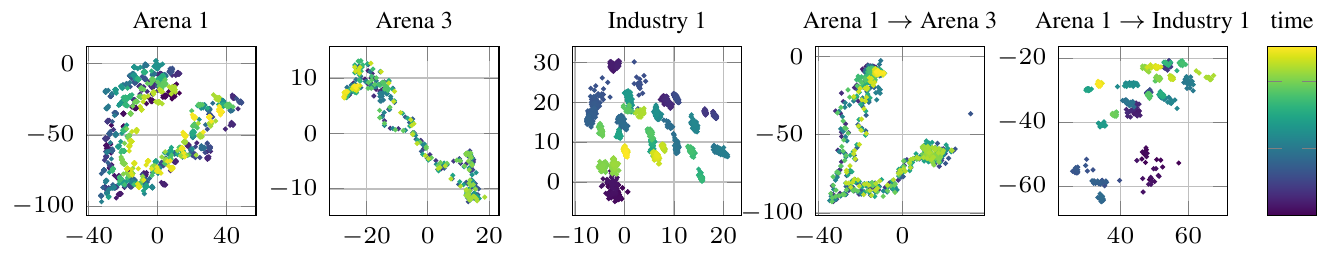}
   \vspace{-0.25cm}
   \caption{Zero-shot performance of channel charting. CT \& TW are two metrics to quantify the goodness of produced CCs~\citep{studer2018channel}. (CT, TW) tuples respectively from left to right: (0.98,0.96), (0.94,0.93), (0.97, 0.94), (0.89, 0.91), (0.96, 0.92). Colourmap denotes time evolution.}
   \label{fig:channel_charts}
\end{figure}

\noindent \textbf{7 -- Shift: Distribution shifts make the zero-shot performance of channel charting meaningless.}
Channel charting (CC) has two drawbacks. Fist, CC outputs low dimensional mapping of the environment (2D, cf. Tab.~\ref{tab:model_variants}) that struggles to faithfully embed the global structure of CSI~\citep{karmanov2021wicluster}.
\citep{karmanov2021wicluster} addressed this dimensionality bottleneck by inflating CC to be more AE-like, which makes us question the use of CC (and not AE) in the first place.
Second, CC requires continuity in sampling the physical environment in order to enforce spatial coherence in its 2D mapping.
Together, these two drawbacks severely hamper the zero-shot application of CC on unseen domains.
Concretely, Fig.~\ref{fig:channel_charts} depicts two CCs trained for Arena 1 and Arena 3.
Referring to the physical space Arena 1 and Arena 3 traverse in Fig.~\ref{fig:lmu_experiments}a, we note two trajectories: blue for Arena 1 and red for Arena 3.
It is interesting to note that when feeding Arena 3 data into Arena 1 CC (Arena 1 $\rightarrow$ Arena 3 in Fig.~\ref{fig:channel_charts}), we obtain a CC that largely overlaps with Arena 1 CC despite the completely different physical trajectories in Fig.~\ref{fig:lmu_experiments}a (i.e., blue vs. red).
This exposes a major limitation of CC w.r.t. zero-shot performance on unseen data.

\begin{figure}[h]
    \centering
    \includegraphics{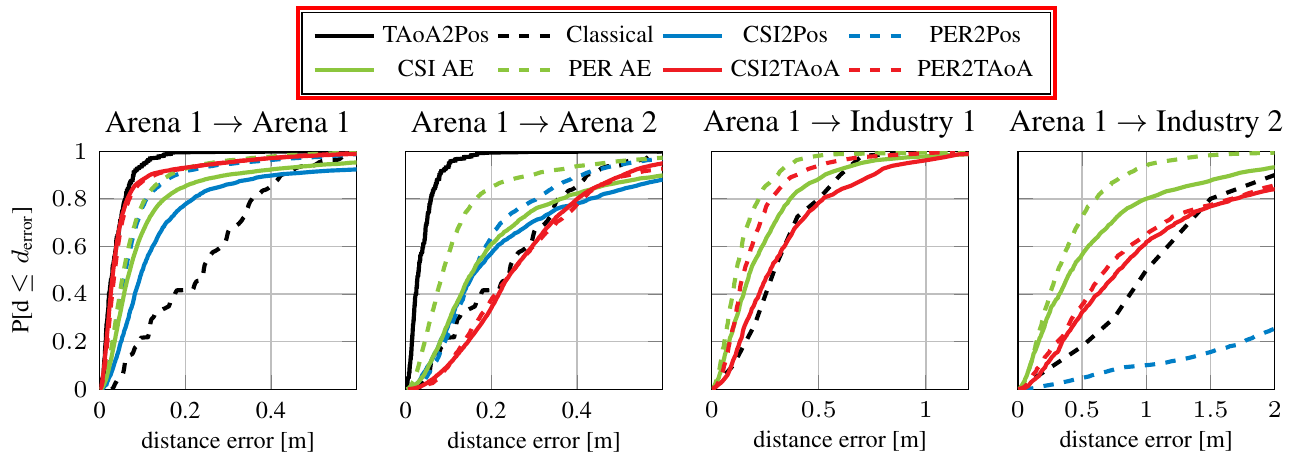}
    \vspace{-0.25cm}
    \caption{Localisation perf. with pretraining on Arena 1 transferred to other domains. Variants xAE and x2TAoA use the frozen pretrained backbones and nonlinear heads trained on 1k samples. The monolithic x2Pos variants use full finetuning on 1k samples.}
        \vspace{-0.25cm}
    \label{fig:frozenBackbone_nonlinearHead_pos_perf}
\end{figure}

\noindent \textbf{8 -- Localisation: Pretraining is a powerful tool for building environment-specific localiser nets.}
The combination of a pretrained backbone and a specialised nonlinear head is a powerful formula for easier deployment of localisation services.
We pretrain backbones for the AE and TAoA configurations on Arena 1.
With 1k samples, we then train nonlinear heads on top of these frozen backbones to specialise for other datasets.
Results in Fig.~\ref{fig:frozenBackbone_nonlinearHead_pos_perf} show that pretraining indeed gives strong performance on other unseen domains.
Specifically, the PER AE configuration outperforms classical by a large margin, while CSI AE, CSI2TAoA, PER2TAoA exhibit competitive performances.
Classical struggles particularly in scattering-rich Industry 2.
Note that in Arena 1 $\rightarrow$ Arena 1 under no distribution shift, all variants outperform Classical.
It is only under distribution shift that we observe clear differentiation in the robustness of the pretrained backbones of model variants.

\subsection{Practical tips}
Based on our analyses, we would recommend the following tips.

\noindent \textbf{1 -- Representation learning is important.} 
Multiple analyses point to qualitative and quantitative differentiation between model variants from~\sysname.
Specifically, moving directly to position and not the latent space results in poor OOD robustness. 
Using a high-dimensional latent space improves robustness to macro and micro distribution shifts, e.g., environment-induced and rich scattering.

\noindent \textbf{2 -- Pretraining is powerful.}
Uniformly across analyses, we found pretraining to be effective for enhancing OOD robustness. 
Further, fine-tuning pretrained representations seems to only require a small  fraction of labels (relative to spatial sampling grid size) for good domain adaptation in RF localiser nets.

\noindent \textbf{3 -- The increased dimensionality of latent space helps.}
Using a latent space with enough dimensionality enhances robustness to distribution shift as well as boosts accuracy. 

\section{Related Work}\label{Sec:related_work}

\noindent \textbf{Learning-based localisation.}
Newer localisation techniques use machine learning.
The rationale is that data-driven learning is able to model and compensate for the sources of error that limit the performance of classic methods.
Machine learning can be applied to both radar techniques~\citep{zhao2018rf,zhao2018through}, or device-based methods~\citep{arnold2019novel,studer2018channel,khatab2017fingerprint}.
Learning-based localisation can either be: supervised~\citep{decurninge2018csi,arnold2019novel}, or unsupervised~\citep{studer2018channel}.

Related works exploit prior floorplan information and feature learning to achieve robustness in RF localisation~\citep{ghazvinian2021modality,kuldeep2021,Kadambi2022,karmanov2021wicluster}. 
Work in~\citep{zanjani2022deep} demonstrates that adding a large metal reflector in an environment causes significant distribution shift that hampers learnt localisation performance.
In contrast to this prior art, out work is the first to: (1) investigate robustness to distribution shift on empirical large-scale data, (2) elucidate how distribution shift impacts learnt localiser variants non-uniformly, and (3) recommend best practices that help enhance robustness.

\noindent \textbf{Benchmarking out of distribution (OOD) robustness.} 
Multiple related works study and benchmark OOD robustness and generalisation as they pertains to mainstream modalities in machine learning~\citep{hendrycks2021many,koh2021wilds,wiles2021}.
Informed and inspired by these works, in this paper we contribute analyses that: (1) compile all RF localiser net variants from prior art, and (2) benchmark their performance from an OOD robustness standpoint.
The latter is crucial for the maturity and deployment of such models in a particularly distribution shift-prone RF environment.

\section{Limitations}

We took first steps towards studying the robustness barrier to real-world deployability of learnt RF localisation.
Robustness to OOD is the number 1 issue outstanding in prior art.
Naturally, there are many more avenues of future investigations.
A few come to mind: (1) broader cross-fertilisation of generalisation methods from mainstream modalities for RF localisation, (2) investigating hardware-induced distribution shift, e.g., antenna configurations, (3) smarter selection criteria of samples to aid distribution shift mitigation, and (4) incorporating sampling in time to further aid robustness.

\section{Conclusion}

In this work, we build a comprehensive framework to analyse and benchmark RF localiser net variants. 
We introduce 5 novel and large-scale datasets curated in industrial premises.
The 5 datasets make possible the empirical study of learning-based RF localisation, and its robustness to macro- and micro-induced distribution shift effects.
Our characterisation shows that localiser net variants without a latent space struggle under distributional shift. 
Our characterisation also shows that there exists a trade-off between accuracy and robustness (including against rich scattering), where AE variants seem to perform especially well. 
We distil our findings into a set of concrete takeaways, a number of practical tips, and open research directions. 
We hope that our benchmarking framework would help foster future research towards realising accurate, rapid, and robust deployments of learnt RF localisation.

\bibliography{conf2023_conference}
\bibliographystyle{conf2023_conference}

\appendix
\newpage
\section{Background on RF Localisation} \label{sec:rf_loc_background}
ToA and AoA estimates are traditionally arrived at using super-resolution methods~\citep{henninger2022probabilistic}. \ac{MLE} then uses ToA and AoA jointly in order to obtain position estimates. 

\noindent \textbf{Positioning model.}
We assume a positioning system comprised of $M$ synchronised locators that listen for user devices.
Each $m$th locator has known 3D position vector $\mathbf{p}_m = [x_m, y_m, z_m]$, and 3D $3 \times 3$ orientation matrix $\mathbf{\Omega}_{m}$. The estimation parameter is the user position $\mathbf{x} = [x, y, z]$. 
$\mathbf{\Omega}_{m}$ is formed by the orthogonal unit vectors corresponding to the $m$th locator's orientation w.r.t. the reference coordinate system.
Each $k$th locator can estimate the user's distance $\hat{d}_m$ that corresponds to its ToA and the impinging  \ac{AoA}. ToA estimates implicitly include an unknown transmit time $\tau$, since user devices are asynchronous w.r.t. the locators.

\noindent \textbf{Joint ToA-AoA MLE.} \label{sec:joint_ml}
Assuming independent ToA and AoA estimates, which can be combined to obtain a more robust joint ToA-AoA position estimate
\begin{align}
(\hat{\mathbf{x}}_{\cap}, \hat{\tau}_{\cap})
& = \argmax_{\mathbf{x}, \tau} \big \{ \sum_{m = 1}^{M} w_{T, m} \cdot \ln \Big \{
\frac{1}{\sqrt{2\pi}} \cdot \exp{ \Big[  
- \frac{(\hat{d}_m - \lVert \mathbf{p}_m - \mathbf{x} \rVert - \tau \cdot c)^{2}}{2\sigma_{m}^{2}}
\Big]} 
\Big\}  \\ 
& +  \sum_{m} \kappa_{m} \hat{\mathbf{u}}_{m}^{\text{T}} \mathbf{\Omega}_{m}^{\text{T}} 
\frac{\mathbf{x} - \mathbf{p}_{m}}{\lVert \mathbf{x} - \mathbf{p}_{m} \rVert} \big \}
\label{eq:joint_ll_fun}
\end{align}
where $\sigma_m^2$ and $w_{T, km}$ are respectively the error variance and optimisation weight that correspond to the $k$th locator, $\tau$ denotes the unknown transmit time, and $c$ the speed of light.
where $\hat{\mathbf{u}}_{m} \in {\mathbb R}^3$ is a unit vector estimate of AoA in the locator's reference frame $\mathbf{\Omega}_{m}$.
The concentration parameters ${\kappa}_m$ reflect the reliability of the angular measurements. 

\subsection{Deep Learning}
Due to the high dimensionality of \ac{RF} data \ac{DL} techniques emerged to close the gap of localization systems in rich scattering environments.
\begin{figure}[H]
    \centering
    \includegraphics{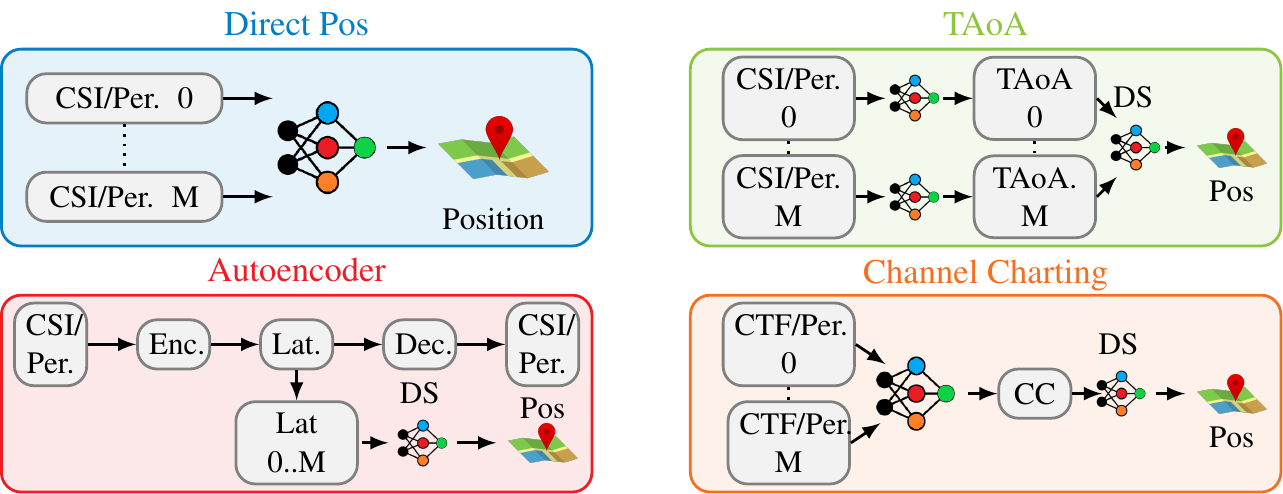}
    \caption{Different procedures to estimate the position from \ac{RF} data.}
    \label{fig:nn_configs}
\end{figure}
Fig.~\ref{fig:nn_configs} shows an extract of the currently implemented NN configurations, which are partly supervised and unsupervised with an additional \ac{DS} task to predict the device position. The methods can be split into different types of prediction challenges. The first proposed techniques tried to estimate directly from the \emph{raw} \ac{CSI} the position directly, expecting the \ac{NN} to capture the channel model and invert it to the position. Another type is predicting the \ac{TAoA} in a local coordinate system to remove any global structure and therefore a robust feature. An alternative to this low dimensional feature is the \ac{AE} which allows to capture the spatial consistency of the channel in the latent space. The currently most favoured approach is \ac{CC} where a low-dimensional channel chart is estimated, by forcing a spatial consistency in the channel chart.
 
\subsection{Supervised Learning}
Supervised learning was the first approach to be tackled, where a huge amount of costly labels were created using advanced \ac{LiDaR} systems. We consider three different strains:

\noindent \textbf{1.) Directly to position from high dimensionality CSI/PER} \newline
Using this high dimensional information the direct mapping,
\begin{equation*} 
\mathcal {C} : \mathbb {C}^{M} \to \mathbb {R} ^{3}
\end{equation*} 
is learned, where fig.~\ref{fig:nn_configs} part one shows this concept. It was shown that this approach generally results in a high dimensionality fingerprinting system. 

\noindent \textbf{2.) Predicting \ac{TAoA}} \newline
Predicting the \ac{TAoA} from the raw information allows to learn a similar mapping
\begin{equation*} 
\mathcal {C} : \mathbb {C}^{M} \to \mathbb {R} ^{3}
\end{equation*} 
but converting the output dimensionality in a local coordinate system independent from other influences. To convert the \ac{TAoA} information into a position a classical weighted system \citep{henninger2022probabilistic} could be used or as shown in fig.~\ref{fig:nn_configs} part two a downstream \ac{NN}.

 \begin{figure}[H]
    \centering
     \includegraphics{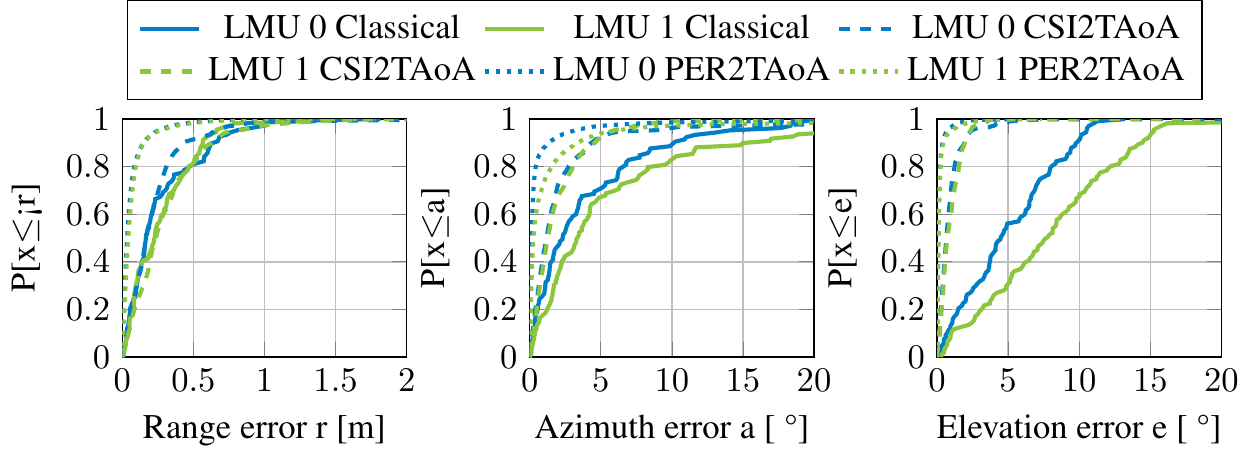}
     \caption{Example prediction of local coordinates using the classical or the proposed \acp{NN}.}
     \label{fig:cdf_range_az_vs_classical}
 \end{figure}
 Fig.~\ref{fig:cdf_range_az_vs_classical} depicts that \ac{NN} can beat due to resilience against hardware impairments (non ideal antenna aperture and oscillators) and rich scattering, where the classical approach crumbles.

\noindent \textbf{3.) Directly to position from TAoA}
An alternative flavour is to predict from the \ac{TAoA} the position. Using multiple locators results in a weighting based on the position system, and thus a form of direct memorization between regions. 

\subsection{Unsupervised Learning}
Due to the costs of creating precise labels as well as updating the labels if the environment changes, multiple  unsupervised approaches hoping to generalize better were proposed. We highlight only two currently most favoured methods.

\subsubsection{Channel Charting}
\ac{CC} exploits the spatial consistency of the channel by first normalizing it with the path-loss via
\begin{equation*} 
\tilde {\mathbf {H}} = \frac {B^{\beta -1}}{\|\bar {\mathbf {H}} \|^\beta _{F}} \bar {\mathbf {H}} 
\end{equation*}
where $B$ is the number of receive antennas, $\beta$ the path-loss exponent and $||_F$ the forbenius norm, respectively. As this technique relies on distiling the spatial consistency by a (hopefully) loss-less dimensionality reduction (e.g. averaging the number of subcarriers), the channel chart can be learned by the mapping
\begin{equation*}
\mathcal {C} : \mathbb {C}^{M'} \to \mathbb {R} ^{D},
\end{equation*} 
where $D$ is typically a low dimensional vector (normally two). This channel chart is learned using the triplet loss. 

As this method is fully unsupervised an the mapping is not corresponding to the actual ground-truth three different metrics were introduced:
(i) \ac{CT} measuring for the K neighbours in the original space the point-wise continuity, e.g. the points are following the same order in the original space as well as the latent space (ii) \ac{TW} measuring he point-wise trustworthiness of the point surrounding in the latent space, e.g. the false points close to the original data point reduces this metric. For more details we refer to \citep{studer2018channel}

\subsubsection{Autoencoder}
The basic idea of an AE is to learn two functions, an encoder C and a decoder $C^{-1}$, so that the average approximation error 
\begin{equation}
E = \frac {1}{N}\sum _{n=1}^{N}\|\mathbf {f} _{n} - \mathcal {C} ^{-1}(\mathcal {C} (\mathbf {f} _{n}))\|_{2}^{2}
\end{equation}
for a set of vectors ${fn}Nn=1$ is minimal. The hope is that the AE creates a low dimensional representation, capturing  the essential components of the channel. These components are later exploited for a downstream task.

\newpage
\section{DL Models} \label{app:dl}

\begin{table}
\centering
\tiny
\caption{TAoA2Pos}
\begin{tabular}{c|c|c|c}
Parameter  & Type & Choices & Final \\ \hline 
epochs & choice & [20, 30, 40] & 20 \\ 
lr & choice & [0.0001, 0.001, 0.01] & 0.0001 \\ 
momentum & uniform & [0.8, 1] & 0.8392178101801823 \\ 
step size & choice & [1, 2, 10] & 10 \\ 
gamma & uniform & [0.4, 0.8] & 0.41178097722520307 \\ 
act func & choice & ['ReLU', 'LeakyReLU', 'Sigmoid', 'Tanh', 'Softplus'] & ReLU \\ 
last act func & choice & ['Sigmoid'] & Sigmoid \\ 
optimizer & choice & ['SGD', 'ADAM'] & SGD \\ 
loss func & choice & ['MSE', 'L1'] & MSE \\ 
lin1 size & choice & [32, 64, 128, 256, 512] & 512 \\ 
lin2 size & choice & [32, 64, 128, 256, 512] & 128 \\ 
lin3 size & choice & [32, 64, 128, 256, 512] & 128 \\ 
\end{tabular}
\end{table}

\begin{figure}
   \centering
    \includegraphics[width=1\textwidth]{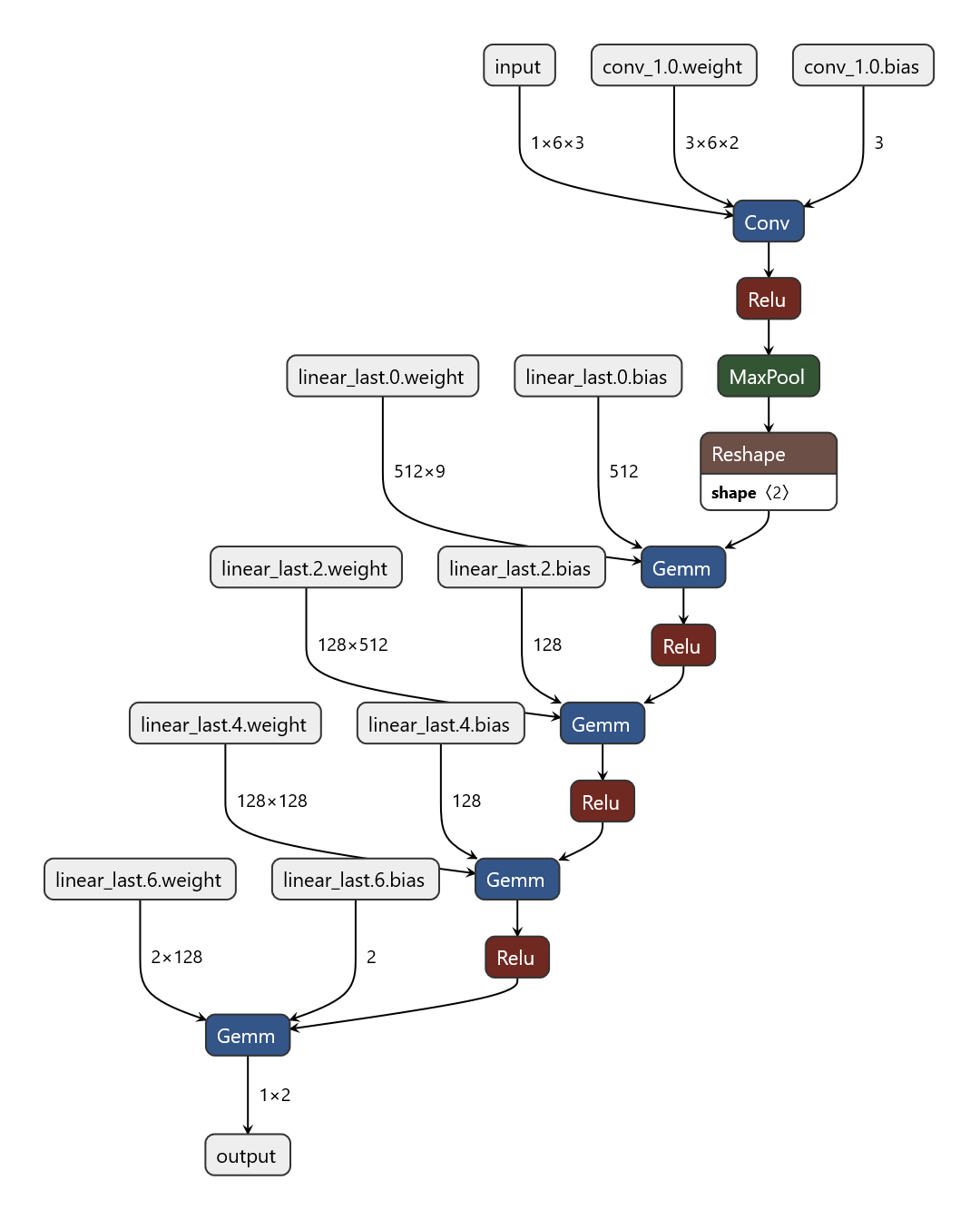}
   \caption{Structure TAoA2Pos}
   \label{fig:taoa2pos}
\end{figure}

\begin{table}
\caption{CSI2Pos}
\tiny
\centering
\begin{tabular}{c|c|c|c}
Parameter  & Type & Choices & Final \\ \hline 
epochs & choice & [20, 30, 40] & 20 \\ 
lr & choice & [0.0001, 0.001, 0.01] & 0.0001 \\ 
momentum & uniform & [0.8, 1] & 0.8853297100815697 \\ 
step size & choice & [1, 2, 10] & 10 \\ 
gamma & uniform & [0.4, 0.8] & 0.43883862858760825 \\ 
act func & choice & ['ReLU', 'LeakyReLU', 'Sigmoid', 'Tanh', 'Softplus'] & ReLU \\ 
last act func & choice & ['Sigmoid'] & Sigmoid \\ 
optimizer & choice & ['SGD', 'ADAM'] & SGD \\ 
loss func & choice & ['MSE', 'L1'] & L1 \\ 
branch1 & choice & [0, 1] & 1 \\ 
branch2 & choice & [0, 1] & 1 \\ 
c1 size & choice & [8, 16, 32, 64] & 8 \\ 
c2 size & choice & [8, 16, 32, 64] & 16 \\ 
c3 size & choice & [8, 16, 32, 64] & 64 \\ 
c4 size & choice & [8, 16, 32, 64] & 8 \\ 
c5 size & choice & [8, 16, 32, 64] & 64 \\ 
c6 size & choice & [8, 16, 32, 64] & 32 \\ 
c7 size & choice & [8, 16, 32, 64] & 16 \\ 
c8 size & choice & [8, 16, 32, 64] & 8 \\ 
c9 size & choice & [8, 16, 32, 64] & 16 \\ 
c10 size & choice & [8, 16, 32, 64] & 16 \\ 
k2 size & choice & [1, 2, 3, 4, 5, 6] & 4 \\ 
k4 size & choice & [1, 2, 3, 4, 5, 6] & 3 \\ 
k6 size & choice & [1, 2, 3, 4, 5, 6] & 4 \\ 
k8 size & choice & [1, 2, 3, 4, 5, 6] & 3 \\ 
k10 size & choice & [1, 2, 3, 4, 5, 6] & 5 \\ 
k12 size & choice & [1, 2, 3, 4, 5, 6] & 2 \\ 
k14 size & choice & [1, 2, 3, 4, 5, 6] & 3 \\ 
k16 size & choice & [1, 2, 3, 4, 5, 6] & 6 \\ 
k18 size & choice & [1, 2, 3, 4, 5, 6] & 6 \\ 
k20 size & choice & [1, 2, 3, 4, 5, 6] & 3 \\ 
s2 size & choice & [1, 2, 3, 4, 5] & 2 \\ 
s4 size & choice & [1, 2, 3, 4, 5, 6] & 1 \\ 
s6 size & choice & [1, 2, 3, 4, 5, 6] & 1 \\ 
s8 size & choice & [1, 2, 3, 4, 5, 6] & 3 \\ 
s10 size & choice & [1, 2, 3, 4, 5, 6] & 2 \\ 
s12 size & choice & [1, 2, 3, 4, 5] & 4 \\ 
s14 size & choice & [1, 2, 3, 4, 5, 6] & 1 \\ 
s16 size & choice & [1, 2, 3, 4, 5, 6] & 5 \\ 
s18 size & choice & [1, 2, 3, 4, 5, 6] & 5 \\ 
s20 size & choice & [1, 2, 3, 4, 5, 6] & 1 \\ 
m2 size & choice & [1, 2, 3, 4, 5, 6] & 6 \\ 
m4 size & choice & [1, 2, 3, 4, 5, 6] & 3 \\ 
m6 size & choice & [1, 2, 3, 4, 5, 6] & 1 \\ 
m8 size & choice & [1, 2, 3, 4, 5, 6] & 2 \\ 
m10 size & choice & [1, 2, 3, 4, 5, 6] & 5 \\ 
m12 size & choice & [1, 2, 3, 4, 5] & 1 \\ 
m14 size & choice & [1, 2, 3, 4, 5, 6] & 6 \\ 
m16 size & choice & [1, 2, 3, 4, 5, 6] & 4 \\ 
m18 size & choice & [1, 2, 3, 4, 5, 6] & 5 \\ 
m20 size & choice & [1, 2, 3, 4, 5, 6] & 4 \\ 
\end{tabular}
\end{table}

\begin{figure}
   \centering
    \includegraphics[width=1\textwidth]{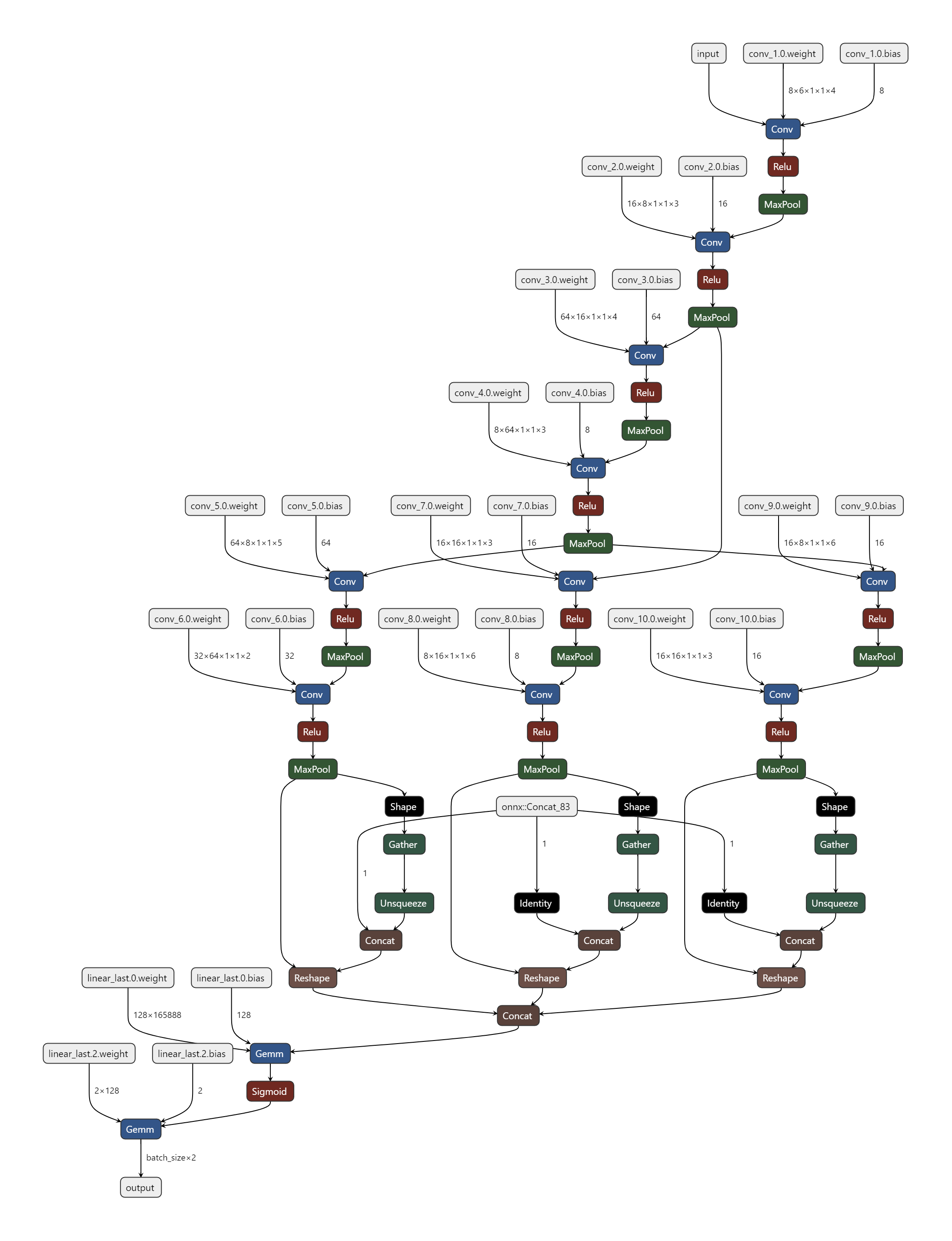}
   \caption{Structure CSI2Pos}
   \label{fig:csi2pos}
\end{figure}

\begin{table}
\caption{CSI2TAoA}
\tiny
\centering
\begin{tabular}{c|c|c|c}
Parameter  & Type & Choices & Final \\ \hline 
epochs & choice & [20, 30, 40] & 30 \\ 
lr & choice & [0.0001, 0.001, 0.01] & 0.0001 \\ 
momentum & uniform & [0.8, 1] & 0.960095906476885 \\ 
step size & choice & [1, 2, 10] & 10 \\ 
gamma & uniform & [0.4, 0.8] & 0.7152406448940722 \\ 
act func & choice & ['ReLU', 'LeakyReLU', 'Sigmoid', 'Tanh', 'Softplus'] & Tanh \\ 
last act func & choice & ['Sigmoid'] & Sigmoid \\ 
optimizer & choice & ['SGD', 'ADAM'] & ADAM \\ 
loss func & choice & ['MSE', 'L1'] & L1 \\ 
branch1 & choice & [0, 1] & 1 \\ 
branch2 & choice & [0, 1] & 0 \\ 
c1 size & choice & [8, 16, 32, 64] & 8 \\ 
c2 size & choice & [8, 16, 32, 64] & 32 \\ 
c3 size & choice & [8, 16, 32, 64] & 8 \\ 
c4 size & choice & [8, 16, 32, 64] & 16 \\ 
c5 size & choice & [8, 16, 32, 64] & 8 \\ 
c6 size & choice & [8, 16, 32, 64] & 64 \\ 
c7 size & choice & [8, 16, 32, 64] & 16 \\ 
c8 size & choice & [8, 16, 32, 64] & 32 \\ 
c9 size & choice & [8, 16, 32, 64] & 64 \\ 
c10 size & choice & [8, 16, 32, 64] & 16 \\ 
k1 size & choice & [1] & 1 \\ 
k2 size & choice & [1, 2, 3, 4, 5, 6] & 1 \\ 
k3 size & choice & [1] & 1 \\ 
k4 size & choice & [1, 2, 3, 4, 5, 6] & 1 \\ 
k5 size & choice & [1] & 1 \\ 
k6 size & choice & [1, 2, 3, 4, 5, 6] & 3 \\ 
k7 size & choice & [1] & 1 \\ 
k8 size & choice & [1, 2, 3, 4, 5, 6] & 6 \\ 
k9 size & choice & [1] & 1 \\ 
k10 size & choice & [1, 2, 3, 4, 5, 6] & 2 \\ 
k11 size & choice & [1] & 1 \\ 
k12 size & choice & [1, 2, 3, 4, 5, 6] & 2 \\ 
k13 size & choice & [1] & 1 \\ 
k14 size & choice & [1, 2, 3, 4, 5, 6] & 3 \\ 
k15 size & choice & [1] & 1 \\ 
k16 size & choice & [1, 2, 3, 4, 5, 6] & 3 \\ 
k17 size & choice & [1] & 1 \\ 
k18 size & choice & [1, 2, 3, 4, 5, 6] & 5 \\ 
k19 size & choice & [1] & 1 \\ 
k20 size & choice & [1, 2, 3, 4, 5, 6] & 2 \\ 
s1 size & choice & [1] & 1 \\ 
s2 size & choice & [1, 2, 3, 4, 5] & 1 \\ 
s3 size & choice & [1] & 1 \\ 
s4 size & choice & [1, 2, 3, 4, 5, 6] & 2 \\ 
s5 size & choice & [1] & 1 \\ 
s6 size & choice & [1, 2, 3, 4, 5, 6] & 4 \\ 
s7 size & choice & [1] & 1 \\ 
s8 size & choice & [1, 2, 3, 4, 5, 6] & 6 \\ 
s9 size & choice & [1] & 1 \\ 
s10 size & choice & [1, 2, 3, 4, 5, 6] & 3 \\ 
s11 size & choice & [1] & 1 \\ 
s12 size & choice & [1, 2, 3, 4, 5] & 1 \\ 
s13 size & choice & [1] & 1 \\ 
s14 size & choice & [1, 2, 3, 4, 5, 6] & 3 \\ 
s15 size & choice & [1] & 1 \\ 
s16 size & choice & [1, 2, 3, 4, 5, 6] & 6 \\ 
s17 size & choice & [1] & 1 \\ 
s18 size & choice & [1, 2, 3, 4, 5, 6] & 3 \\ 
s19 size & choice & [1] & 1 \\ 
s20 size & choice & [1, 2, 3, 4, 5, 6] & 3 \\ 
m1 size & choice & [1] & 1 \\ 
m2 size & choice & [1, 2, 3, 4, 5, 6] & 4 \\ 
m3 size & choice & [1] & 1 \\ 
m4 size & choice & [1, 2, 3, 4, 5, 6] & 4 \\ 
m5 size & choice & [1] & 1 \\ 
m6 size & choice & [1, 2, 3, 4, 5, 6] & 2 \\ 
m7 size & choice & [1] & 1 \\ 
m8 size & choice & [1, 2, 3, 4, 5, 6] & 1 \\ 
m9 size & choice & [1] & 1 \\ 
m10 size & choice & [1, 2, 3, 4, 5, 6] & 4 \\ 
m11 size & choice & [1] & 1 \\ 
m12 size & choice & [1, 2, 3, 4, 5] & 3 \\ 
m13 size & choice & [1] & 1 \\ 
m14 size & choice & [1, 2, 3, 4, 5, 6] & 2 \\ 
m15 size & choice & [1] & 1 \\ 
m16 size & choice & [1, 2, 3, 4, 5, 6] & 6 \\ 
m17 size & choice & [1] & 1 \\ 
m18 size & choice & [1, 2, 3, 4, 5, 6] & 5 \\ 
m19 size & choice & [1] & 1 \\ 
m20 size & choice & [1, 2, 3, 4, 5, 6] & 6 \\ 
\end{tabular}
\end{table}

\begin{figure}
   \centering
    \includegraphics[width=0.8\textwidth]{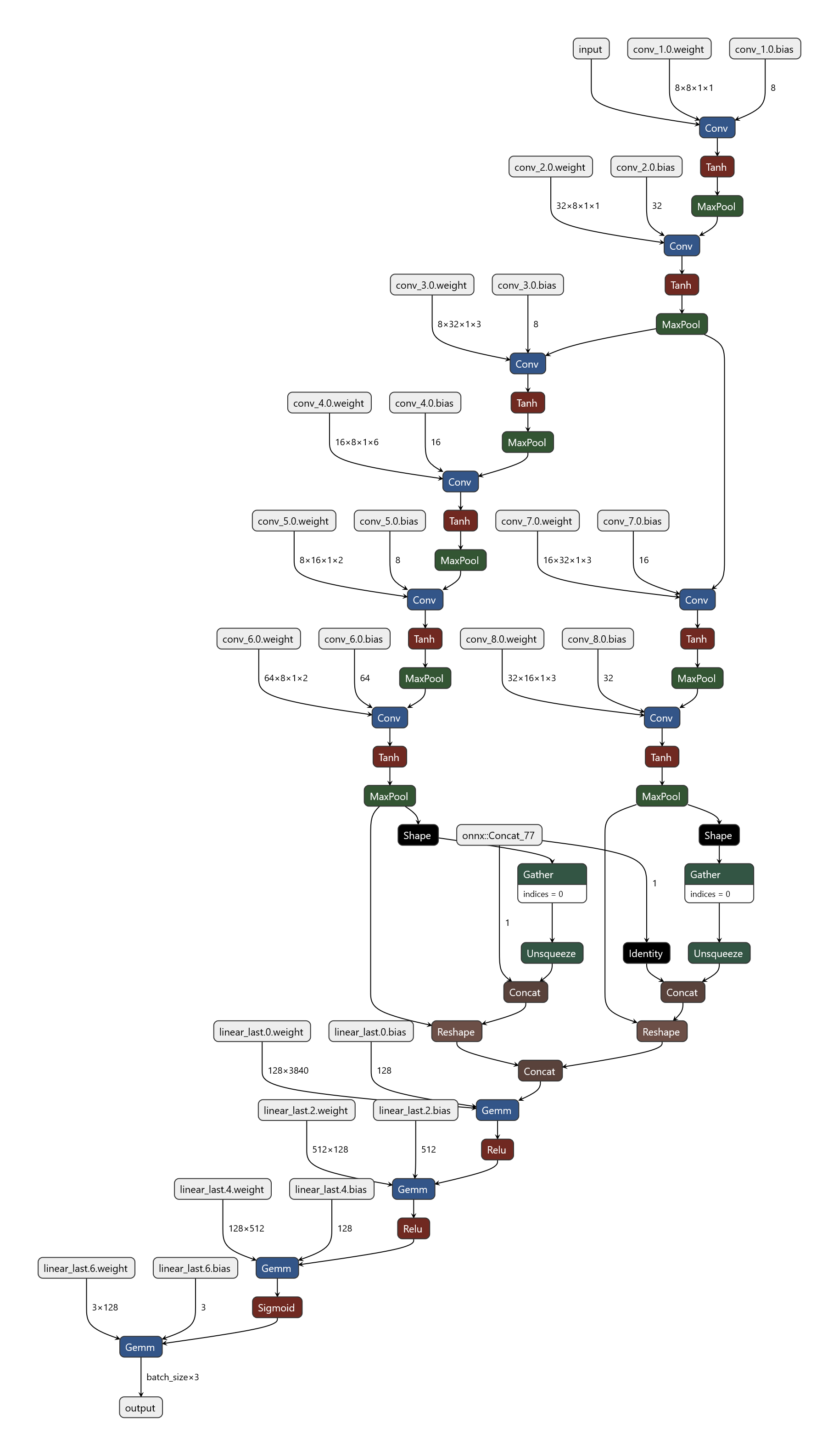}
   \caption{Structure CSI2TAoA}
   \label{fig:csi2taoa}
\end{figure}

\begin{table}
\caption{CSI AE}
\tiny
\centering
\begin{tabular}{c|c|c|c}
Parameter  & Type & Choices & Final \\ \hline 
epochs & choice & [20, 30, 40] & 20 \\ 
lr & choice & [0.0001, 0.001, 0.01] & 0.0001 \\ 
momentum & uniform & [0.8, 1] & 0.9363040304887302 \\ 
step size & choice & [1, 2, 10] & 2 \\ 
gamma & uniform & [0.4, 0.8] & 0.44213564205749145 \\ 
act func & choice & ['ReLU', 'LeakyReLU', 'Sigmoid', 'Tanh', 'Softplus'] & ReLU \\ 
last act func & choice & ['Sigmoid'] & Sigmoid \\ 
optimizer & choice & ['SGD', 'ADAM'] & ADAM \\ 
loss func & choice & ['MSE', 'L1', 'NMSE', 'BCE'] & L1 \\ 
lin1 size & choice & [32, 64, 128, 256, 512] & 256 \\ 
lin2 size & choice & [32, 64, 128, 256, 512] & 64 \\ 
lin3 size & choice & [32, 64, 128, 256, 512] & 128 \\ 
lin4 size & choice & [32, 64, 128, 256, 512] & 128 \\ 
feat dim & choice & [128, 256, 512, 1024] & 256 \\ 
\end{tabular}
\end{table}

\begin{figure}
   \centering
    \includegraphics[width=1\textwidth]{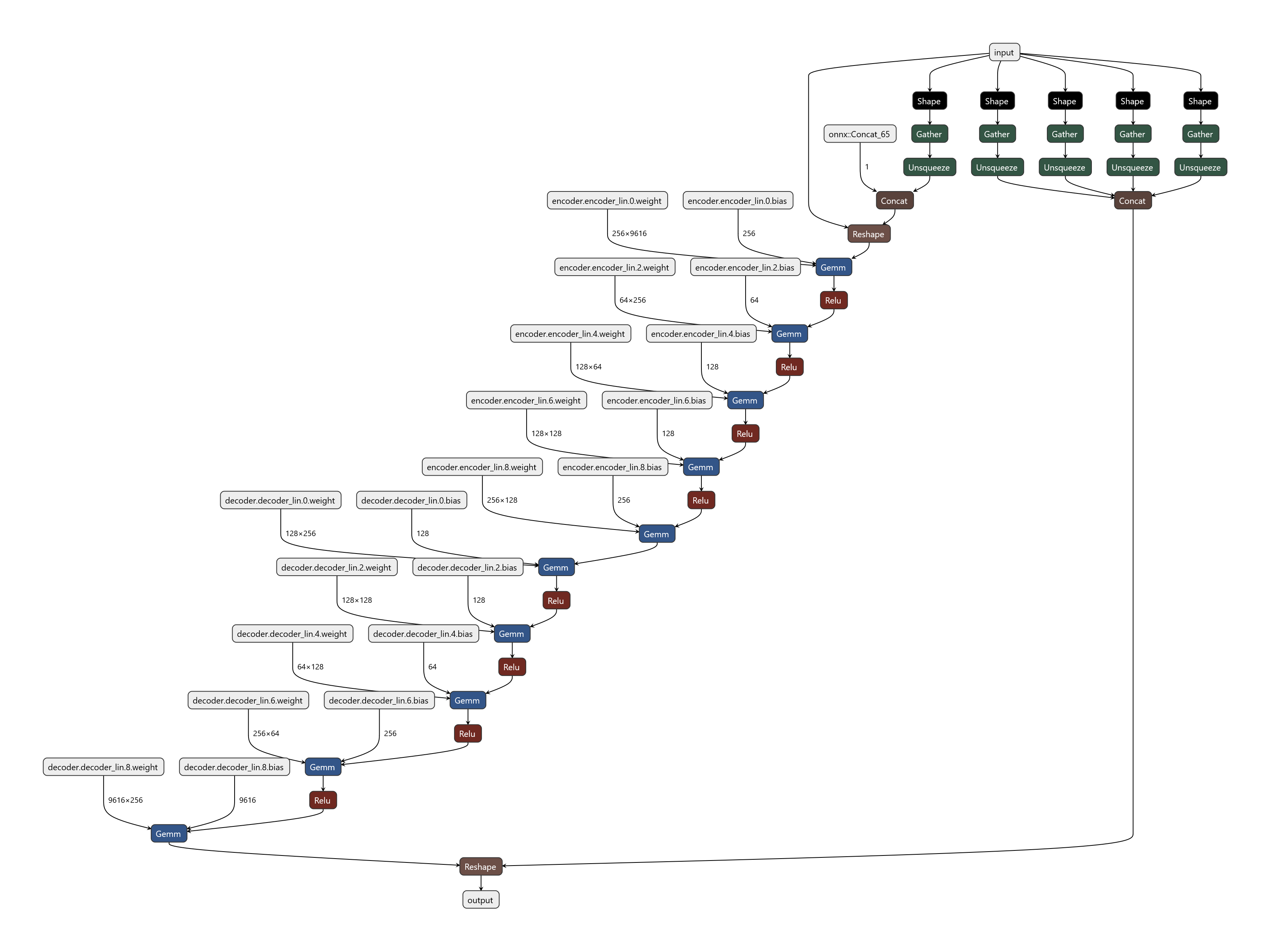}
   \caption{Structure CSI AE}
   \label{fig:csi_ae}
\end{figure}

\begin{table}
\caption{PER2Pos}
\tiny
\centering
\begin{tabular}{c|c|c|c}
Parameter  & Type & Choices & Final \\ \hline 
epochs & choice & [20, 30, 40] & 40 \\ 
lr & choice & [0.0001, 0.001, 0.01] & 0.001 \\ 
momentum & uniform & [0.8, 1] & 0.9997778126234411 \\ 
step size & choice & [1, 2, 10] & 1 \\ 
gamma & uniform & [0.4, 0.8] & 0.560402530889289 \\ 
act func & choice & ['ReLU', 'LeakyReLU', 'Sigmoid', 'Tanh', 'Softplus'] & ReLU \\ 
last act func & choice & ['Sigmoid'] & Sigmoid \\ 
optimizer & choice & ['SGD', 'ADAM'] & ADAM \\ 
loss func & choice & ['MSE', 'L1'] & MSE \\ 
branch1 & choice & [0, 1] & 1 \\ 
branch2 & choice & [0, 1] & 0 \\ 
c1 size & choice & [8, 16, 32, 64] & 64 \\ 
c2 size & choice & [8, 16, 32, 64] & 32 \\ 
c3 size & choice & [8, 16, 32, 64] & 16 \\ 
c4 size & choice & [8, 16, 32, 64] & 8 \\ 
c5 size & choice & [8, 16, 32, 64] & 32 \\ 
c6 size & choice & [8, 16, 32, 64] & 64 \\ 
c7 size & choice & [8, 16, 32, 64] & 8 \\ 
c8 size & choice & [8, 16, 32, 64] & 32 \\ 
c9 size & choice & [8, 16, 32, 64] & 8 \\ 
c10 size & choice & [8, 16, 32, 64] & 8 \\ 
k1 size & choice & [1, 2, 4] & 1 \\ 
k2 size & choice & [1, 2, 4] & 1 \\ 
k3 size & choice & [1, 2] & 1 \\ 
k4 size & choice & [1, 2, 4] & 4 \\ 
k5 size & choice & [1, 2] & 2 \\ 
k6 size & choice & [1, 2, 4] & 4 \\ 
k7 size & choice & [1, 2] & 2 \\ 
k8 size & choice & [1, 2, 4] & 1 \\ 
k9 size & choice & [1, 2] & 1 \\ 
k10 size & choice & [1, 2, 4] & 1 \\ 
k11 size & choice & [1, 2] & 2 \\ 
k12 size & choice & [1, 2, 4] & 4 \\ 
k13 size & choice & [1, 2] & 2 \\ 
k14 size & choice & [1, 2, 4] & 2 \\ 
k15 size & choice & [1, 2] & 2 \\ 
k16 size & choice & [1, 2, 4] & 1 \\ 
k17 size & choice & [1, 2] & 1 \\ 
k18 size & choice & [1, 2, 4] & 2 \\ 
k19 size & choice & [1, 2] & 2 \\ 
k20 size & choice & [1, 2, 4] & 4 \\ 
s1 size & choice & [1, 2] & 4 \\ 
s2 size & choice & [1, 2, 3, 4, 5] & 2 \\ 
s3 size & choice & [1, 2] & 4 \\ 
s4 size & choice & [1, 2, 4] & 1 \\ 
s5 size & choice & [1, 2] & 4 \\ 
s6 size & choice & [1, 2, 4] & 2 \\ 
s7 size & choice & [1, 2] & 4 \\ 
s8 size & choice & [1, 2, 4] & 4 \\ 
s9 size & choice & [1, 2] & 1 \\ 
s10 size & choice & [1, 2, 4] & 4 \\ 
s11 size & choice & [1, 2] & 1 \\ 
s12 size & choice & [1, 2, 3, 4, 5] & 1 \\ 
s13 size & choice & [1, 2] & 2 \\ 
s14 size & choice & [1, 2, 4] & 4 \\ 
s15 size & choice & [1, 2] & 1 \\ 
s16 size & choice & [1, 2, 4] & 1 \\ 
s17 size & choice & [1, 2] & 2 \\ 
s18 size & choice & [1, 2, 4] & 4 \\ 
s19 size & choice & [1, 2] & 8 \\ 
s20 size & choice & [1, 2, 4] & 8 \\ 
m1 size & choice & [1, 2] & 2 \\ 
m2 size & choice & [1, 2, 4] & 2 \\ 
m3 size & choice & [1, 2] & 1 \\ 
m4 size & choice & [1, 2, 4] & 4 \\ 
m5 size & choice & [1, 2] & 1 \\ 
m6 size & choice & [1, 2, 4] & 2 \\ 
m7 size & choice & [1, 2] & 1 \\ 
m8 size & choice & [1, 2, 4] & 1 \\ 
m9 size & choice & [1, 2] & 1 \\ 
m10 size & choice & [1, 2, 4] & 2 \\ 
m11 size & choice & [1, 2] & 1 \\ 
m12 size & choice & [1, 2, 3, 4, 5] & 2 \\ 
m13 size & choice & [1, 2] & 1 \\ 
m14 size & choice & [1, 2, 4] & 2 \\ 
m15 size & choice & [1, 2] & 1 \\ 
m16 size & choice & [1, 2, 4] & 2 \\ 
m17 size & choice & [1, 2] & 1 \\ 
m18 size & choice & [1, 2, 4] & 4 \\ 
m19 size & choice & [1, 2] & 2 \\ 
m20 size & choice & [1, 2, 4] & 4 \\ 
\end{tabular}
\end{table}

\begin{figure}
   \centering
    \includegraphics[width=0.8\textwidth]{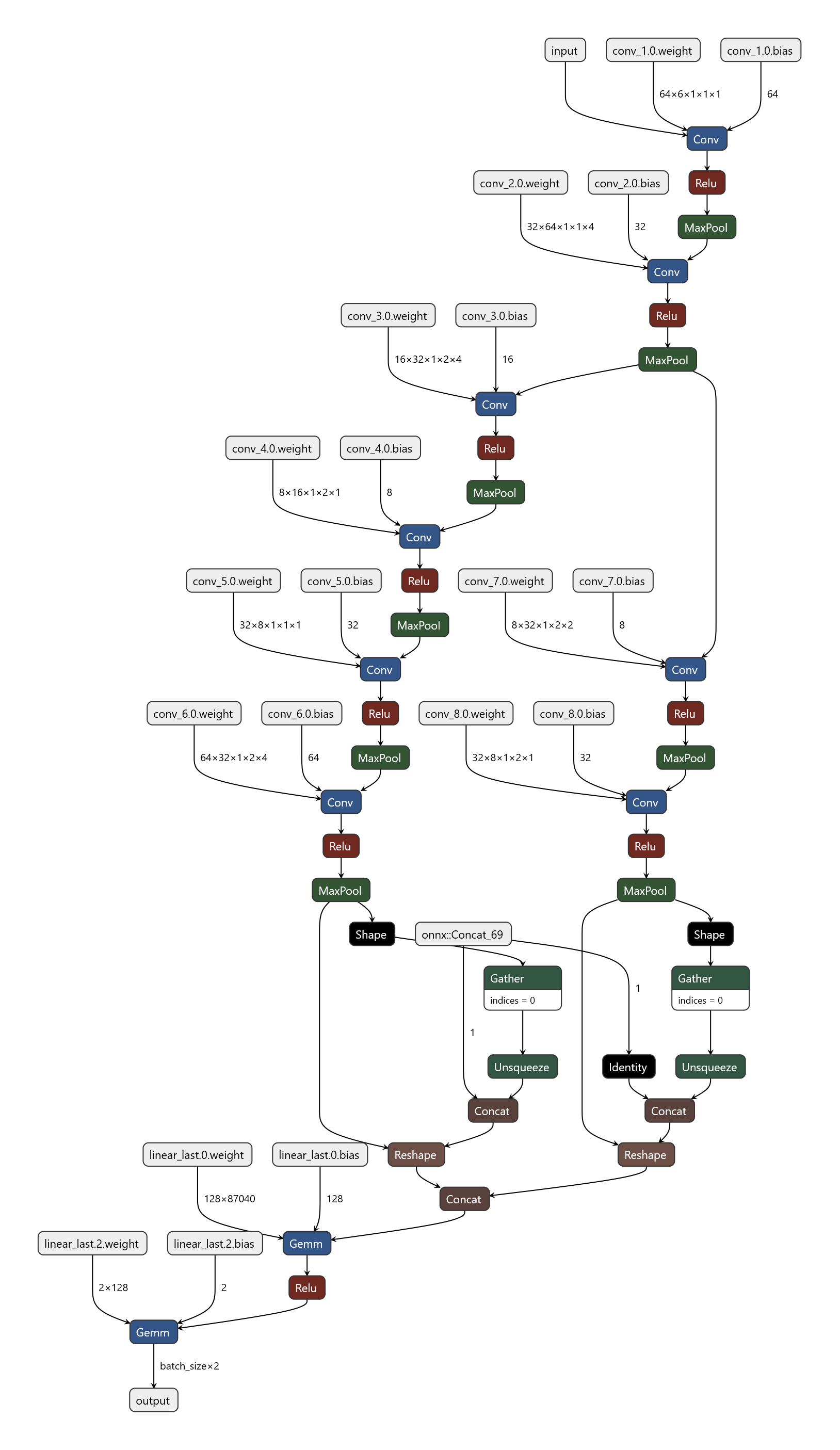}
   \caption{Structure Per2Pos}
   \label{fig:Per2pos}
\end{figure}

\begin{table}
\tiny
\centering
\caption{PER2TAoA}
\begin{tabular}{c|c|c|c}
Parameter  & Type & Choices & Final \\ \hline 
epochs & choice & [20, 30, 40] & 30 \\ 
lr & choice & [0.0001, 0.001, 0.01] & 0.0001 \\ 
momentum & uniform & [0.8, 1] & 0.8999771834863867 \\ 
step size & choice & [1, 2, 10] & 2 \\ 
gamma & uniform & [0.4, 0.8] & 0.7350887010069965 \\ 
act func & choice & ['ReLU', 'LeakyReLU', 'Sigmoid', 'Tanh', 'Softplus'] & Softplus \\ 
last act func & choice & ['Sigmoid'] & Sigmoid \\ 
optimizer & choice & ['SGD', 'ADAM'] & SGD \\ 
loss func & choice & ['MSE', 'L1'] & L1 \\ 
branch1 & choice & [0, 1] & 0 \\ 
branch2 & choice & [0, 1] & 0 \\ 
c1 size & choice & [8, 16, 32, 64] & 16 \\ 
c2 size & choice & [8, 16, 32, 64] & 16 \\ 
c3 size & choice & [8, 16, 32, 64] & 8 \\ 
c4 size & choice & [8, 16, 32, 64] & 16 \\ 
c5 size & choice & [8, 16, 32, 64] & 64 \\ 
c6 size & choice & [8, 16, 32, 64] & 16 \\ 
c7 size & choice & [8, 16, 32, 64] & 16 \\ 
c8 size & choice & [8, 16, 32, 64] & 16 \\ 
c9 size & choice & [8, 16, 32, 64] & 16 \\ 
c10 size & choice & [8, 16, 32, 64] & 8 \\ 
k1 size & choice & [1, 2, 4] & 2 \\ 
k2 size & choice & [1, 2, 4] & 1 \\ 
k3 size & choice & [1, 2] & 1 \\ 
k4 size & choice & [1, 2, 4] & 2 \\ 
k5 size & choice & [1, 2] & 1 \\ 
k6 size & choice & [1, 2, 4] & 1 \\ 
k7 size & choice & [1, 2] & 2 \\ 
k8 size & choice & [1, 2, 4] & 1 \\ 
k9 size & choice & [1, 2] & 2 \\ 
k10 size & choice & [1, 2, 4] & 1 \\ 
k11 size & choice & [1, 2] & 1 \\ 
k12 size & choice & [1, 2, 4] & 2 \\ 
k13 size & choice & [1, 2] & 2 \\ 
k14 size & choice & [1, 2, 4] & 4 \\ 
k15 size & choice & [1, 2] & 1 \\ 
k16 size & choice & [1, 2, 4] & 4 \\ 
k17 size & choice & [1, 2] & 2 \\ 
k18 size & choice & [1, 2, 4] & 2 \\ 
k19 size & choice & [1, 2] & 1 \\ 
k20 size & choice & [1, 2, 4] & 1 \\ 
s1 size & choice & [1, 2] & 2 \\ 
s2 size & choice & [1, 2, 3, 4, 5] & 2 \\ 
s3 size & choice & [1, 2] & 1 \\ 
s4 size & choice & [1, 2, 4] & 1 \\ 
s5 size & choice & [1, 2] & 1 \\ 
s6 size & choice & [1, 2, 4] & 4 \\ 
s7 size & choice & [1, 2] & 2 \\ 
s8 size & choice & [1, 2, 4] & 4 \\ 
s9 size & choice & [1, 2] & 2 \\ 
s10 size & choice & [1, 2, 4] & 1 \\ 
s11 size & choice & [1, 2] & 1 \\ 
s12 size & choice & [1, 2, 3, 4, 5] & 3 \\ 
s13 size & choice & [1, 2] & 2 \\ 
s14 size & choice & [1, 2, 4] & 4 \\ 
s15 size & choice & [1, 2] & 1 \\ 
s16 size & choice & [1, 2, 4] & 1 \\ 
s17 size & choice & [1, 2] & 1 \\ 
s18 size & choice & [1, 2, 4] & 1 \\ 
s19 size & choice & [1, 2] & 4 \\ 
s20 size & choice & [1, 2, 4] & 2 \\ 
m1 size & choice & [1, 2] & 1 \\ 
m2 size & choice & [1, 2, 4] & 4 \\ 
m3 size & choice & [1, 2] & 1 \\ 
m4 size & choice & [1, 2, 4] & 4 \\ 
m5 size & choice & [1, 2] & 2 \\ 
m6 size & choice & [1, 2, 4] & 4 \\ 
m7 size & choice & [1, 2] & 1 \\ 
m8 size & choice & [1, 2, 4] & 4 \\ 
m9 size & choice & [1, 2] & 1 \\ 
m10 size & choice & [1, 2, 4] & 2 \\ 
m11 size & choice & [1, 2] & 2 \\ 
m12 size & choice & [1, 2, 3, 4, 5] & 5 \\ 
m13 size & choice & [1, 2] & 1 \\ 
m14 size & choice & [1, 2, 4] & 2 \\ 
m15 size & choice & [1, 2] & 2 \\ 
m16 size & choice & [1, 2, 4] & 2 \\ 
m17 size & choice & [1, 2] & 2 \\ 
m18 size & choice & [1, 2, 4] & 2 \\ 
m19 size & choice & [1, 2] & 2 \\ 
m20 size & choice & [1, 2, 4] & 2 \\ 
\end{tabular}
\end{table}

\begin{figure}
   \centering
    \includegraphics[width=1\textwidth]{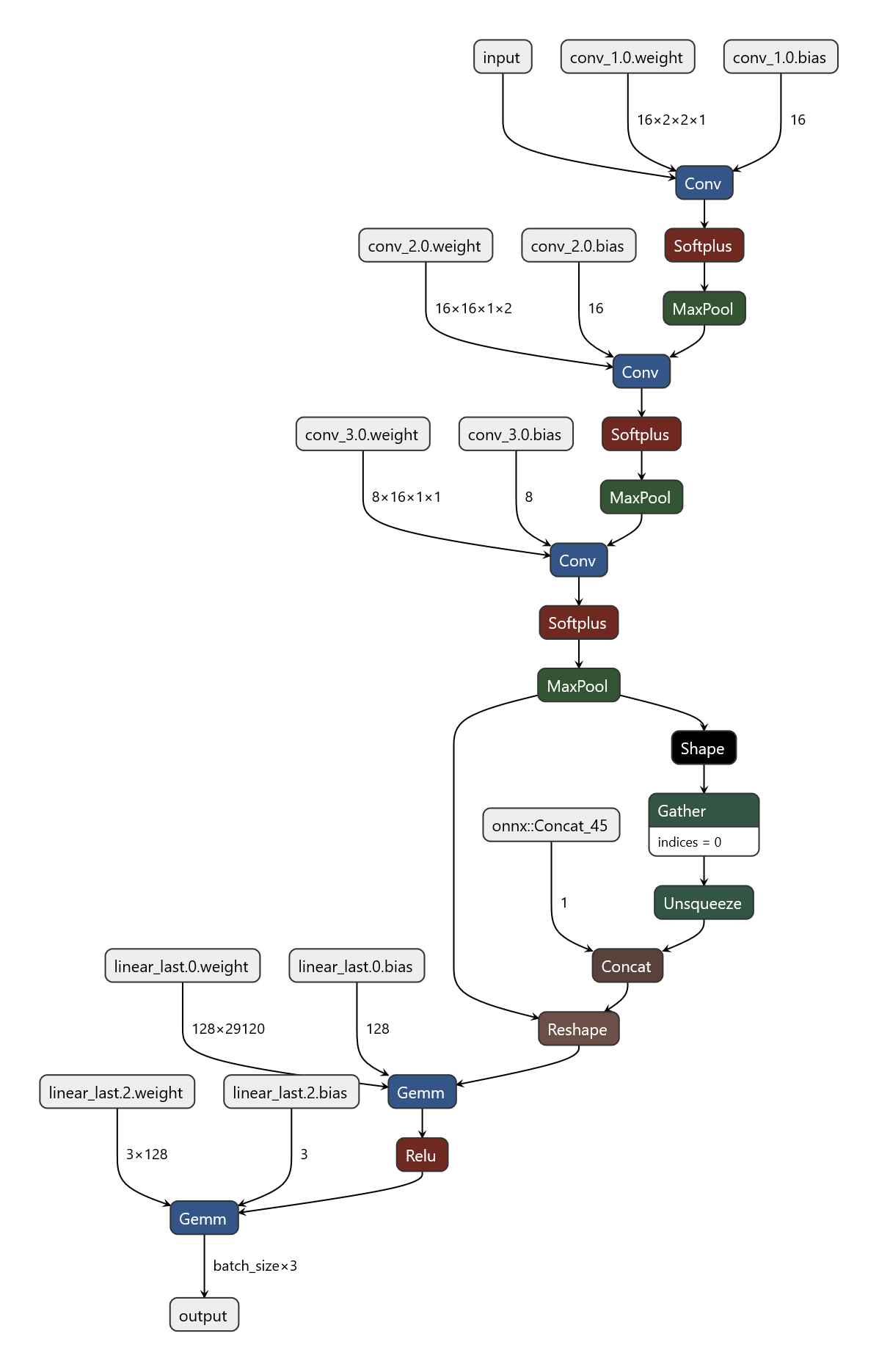}
   \caption{Structure Per2TAoA}
   \label{fig:per2taoa}
\end{figure}

\begin{table}
\tiny
\centering
\caption{PER AE}
\begin{tabular}{c|c|c|c}
Parameter  & Type & Choices & Final \\ \hline 
epochs & choice & [20, 30, 40] & 30 \\ 
lr & choice & [0.0001, 0.001, 0.01] & 0.001 \\ 
momentum & uniform & [0.8, 1] & 0.9593149656519064 \\ 
step size & choice & [1, 2, 10] & 10 \\ 
gamma & uniform & [0.4, 0.8] & 0.870379744744098 \\ 
act func & choice & ['ReLU', 'LeakyReLU', 'Sigmoid', 'Tanh', 'Softplus'] & ReLU \\ 
last act func & choice & ['Sigmoid'] & Sigmoid \\ 
optimizer & choice & ['SGD', 'ADAM'] & ADAM \\ 
loss func & choice & ['MSE', 'L1', 'NMSE', 'BCE'] & BCE \\ 
branch1 & choice & [0, 1] & 1 \\ 
branch2 & choice & [0, 1] & 1 \\ 
c1 size & choice & [8, 16, 32, 64] & 64 \\ 
c2 size & choice & [8, 16, 32, 64] & 32 \\ 
c3 size & choice & [8, 16, 32, 64] & 16 \\ 
c4 size & choice & [1] & 8 \\ 
c5 size & choice & [1] & 32 \\ 
c6 size & choice & [1] & 64 \\ 
c7 size & choice & [1] & 8 \\ 
c8 size & choice & [1] & 32 \\ 
c9 size & choice & [1] & 8 \\ 
c10 size & choice & [1] & 64 \\ 
k1 size & choice & [1] & 1 \\ 
k2 size & choice & [1, 2, 4] & 1 \\ 
k3 size & choice & [1, 2] & 1 \\ 
k4 size & choice & [1, 2, 4] & 4 \\ 
k5 size & choice & [1, 2] & 2 \\ 
k6 size & choice & [1, 2, 4] & 4 \\ 
k7 size & choice & [1] & 2 \\ 
k8 size & choice & [1] & 1 \\ 
k9 size & choice & [1] & 1 \\ 
k10 size & choice & [1] & 1 \\ 
k11 size & choice & [1] & 2 \\ 
k12 size & choice & [1] & 4 \\ 
k13 size & choice & [1] & 2 \\ 
k14 size & choice & [1] & 2 \\ 
k15 size & choice & [1] & 2 \\ 
k16 size & choice & [1] & 1 \\ 
k17 size & choice & [1] & 1 \\ 
k18 size & choice & [1] & 2 \\ 
k19 size & choice & [1] & 2 \\ 
k20 size & choice & [1] & 4 \\ 
s1 size & choice & [1] & 2 \\ 
s2 size & choice & [1, 2, 4] & 2 \\ 
s3 size & choice & [1, 2] & 2 \\ 
s4 size & choice & [1, 2, 4] & 1 \\ 
s5 size & choice & [1, 2] & 1 \\ 
s6 size & choice & [1, 2, 4] & 2 \\ 
s7 size & choice & [1] & 2 \\ 
s8 size & choice & [1] & 2 \\ 
s9 size & choice & [1] & 1 \\ 
s10 size & choice & [1] & 4 \\ 
s11 size & choice & [1] & 1 \\ 
s12 size & choice & [1] & 1 \\ 
s13 size & choice & [1] & 2 \\ 
s14 size & choice & [1] & 4 \\ 
s15 size & choice & [1] & 1 \\ 
s16 size & choice & [1] & 1 \\ 
s17 size & choice & [1] & 2 \\ 
s18 size & choice & [1] & 4 \\ 
s19 size & choice & [1] & 1 \\ 
s20 size & choice & [1] & 1 \\ 
m1 size & choice & [1] & 2 \\ 
m2 size & choice & [1, 2] & 2 \\ 
m3 size & choice & [1] & 1 \\ 
m4 size & choice & [1, 2] & 4 \\ 
m5 size & choice & [1] & 1 \\ 
m6 size & choice & [1, 2] & 2 \\ 
m7 size & choice & [1] & 1 \\ 
m8 size & choice & [1] & 1 \\ 
m9 size & choice & [1] & 1 \\ 
m10 size & choice & [1] & 2 \\ 
m11 size & choice & [1] & 1 \\ 
m12 size & choice & [1] & 2 \\ 
m13 size & choice & [1] & 1 \\ 
m14 size & choice & [1] & 2 \\ 
m15 size & choice & [1] & 1 \\ 
m16 size & choice & [1] & 2 \\ 
m17 size & choice & [1] & 1 \\ 
m18 size & choice & [1] & 4 \\ 
m19 size & choice & [1] & 2 \\ 
m20 size & choice & [1] & 1 \\ 
feat dim & choice & [128, 256, 512, 1024] & 256 \\ 
\end{tabular}
\end{table}

\begin{table}
\tiny
\centering
\caption{PER CC}
\begin{tabular}{c|c|c|c}
Parameter  & Type & Choices & Final \\ \hline 
epochs & choice & [20, 30, 40] & 40 \\ 
lr & choice & [0.0001, 0.001, 0.01] & 0.0001 \\ 
momentum & uniform & [0.8, 1] & 0.8913774253680748 \\ 
step size & choice & [1, 2, 10] & 2 \\ 
gamma & uniform & [0.4, 0.8] & 0.6183381813195497 \\ 
act func & choice & ['ReLU', 'LeakyReLU', 'Sigmoid', 'Tanh', 'Softplus'] & ReLU \\ 
last act func & choice & ['Sigmoid'] & Linear \\ 
optimizer & choice & ['SGD', 'ADAM'] & ADAM \\ 
loss func & choice & ['Triplet'] & Triplet \\ 
branch1 & choice & [0, 1] & 0 \\ 
branch2 & choice & [0, 1] & 1 \\ 
c1 size & choice & [8, 16, 32, 64] & 16 \\ 
c2 size & choice & [8, 16, 32, 64] & 8 \\ 
c3 size & choice & [8, 16, 32, 64] & 16 \\ 
k1 size & choice & [1, 2, 4] & 16 \\ 
k2 size & choice & [1, 2, 4] & 4 \\ 
k3 size & choice & [1, 2, 4] & 4 \\ 
k4 size & choice & [1, 2, 4] & 2 \\ 
k5 size & choice & [1, 2, 4] & 1 \\ 
k6 size & choice & [1, 2, 4] & 4 \\ 
s1 size & choice & [1, 2] & 2 \\ 
s2 size & choice & [1, 2] & 2 \\ 
s3 size & choice & [1, 2] & 2 \\ 
s4 size & choice & [1, 2] & 2 \\ 
s5 size & choice & [1, 2] & 1 \\ 
s6 size & choice & [1, 2] & 2 \\ 
m1 size & choice & [1, 2] & 2 \\ 
m2 size & choice & [1, 2] & 2 \\ 
m3 size & choice & [1, 2] & 2 \\ 
m4 size & choice & [1, 2] & 1 \\ 
m5 size & choice & [1, 2] & 2 \\ 
m6 size & choice & [1, 2] & 2 \\ 
beta & choice & [3] & 3 \\ 
feat dim & choice & [128, 256, 512, 1024] & 256 \\ 
\end{tabular}
\end{table}

\begin{figure}
   \centering
    \includegraphics[width=1\textwidth]{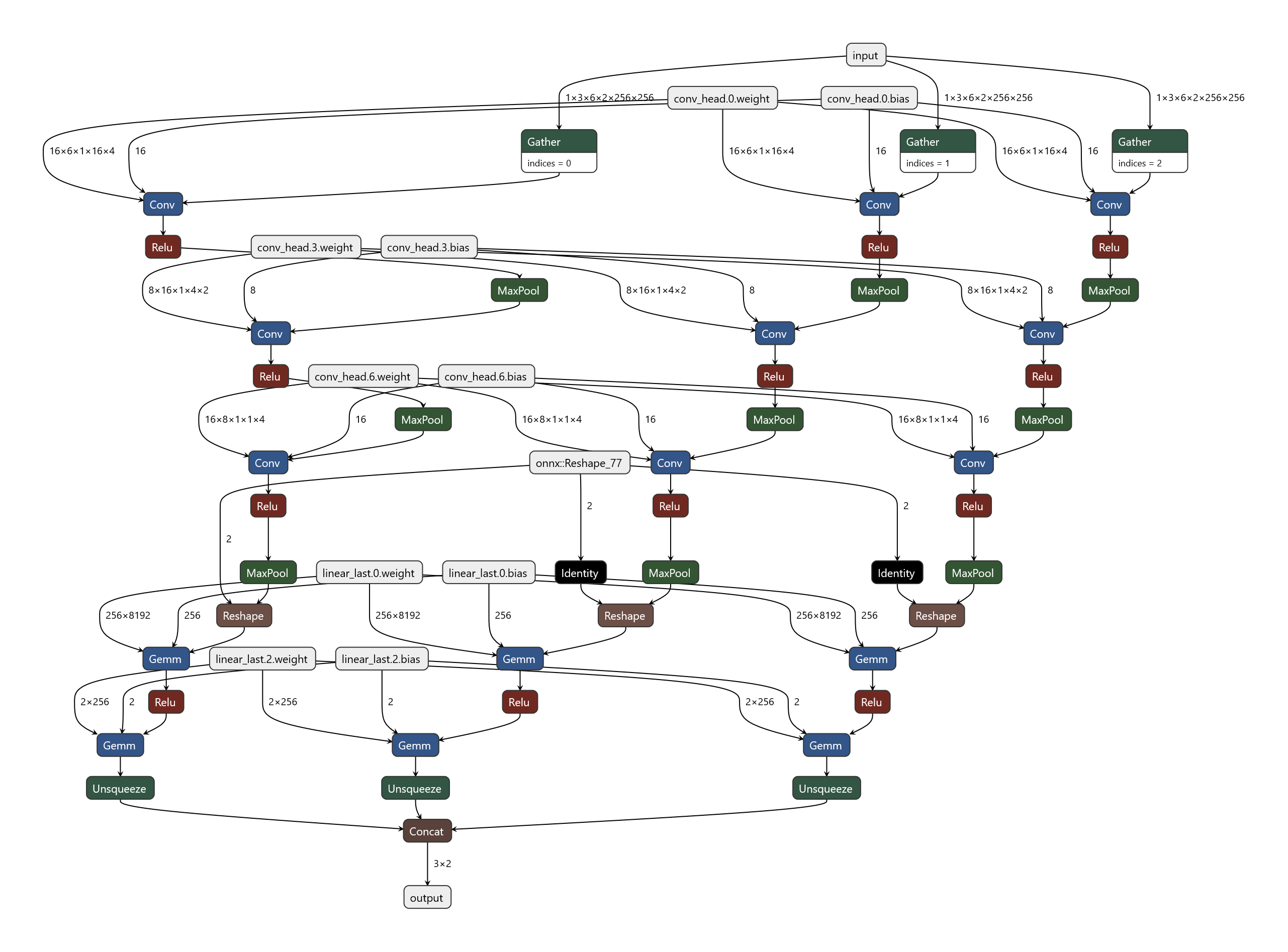}
   \caption{Structure Per CC}
   \label{fig:per_cc}
\end{figure}

\end{document}